\documentclass[10pt,twocolumn,letterpaper]{article}

\usepackage{cvpr}              % To produce the CAMERA-READY version
\usepackage{tabularray}

\definecolor{cvprblue}{rgb}{0.21,0.49,0.74}
\usepackage[pagebackref,breaklinks,colorlinks,allcolors=cvprblue]{hyperref}
\usepackage{kotex}
\usepackage{graphicx}
\usepackage{multirow}
\usepackage{colortbl}
\usepackage{adjustbox}
\usepackage{xcolor}
\usepackage{booktabs}
\usepackage{hhline}
\usepackage{array}
\usepackage{tabularx}
\usepackage{pifont} 

\newcommand{\xmark}{\ding{55}} % X 표시

\def\paperID{2051} % *** Enter the Paper ID here
\def\confName{CVPR}
\def\confYear{2025}

\title{
No Thing, Nothing: Highlighting Safety-Critical Classes\\ for Robust LiDAR Semantic Segmentation in Adverse Weather
}

\author{Junsung Park
\and
Hwijeong Lee
\and
Inha Kang
\and
Hyunjung Shim\\
Korea Advanced Institute of Science and Technology\\
108, Taebong-ro, Seocho-gu, Seoul, Republic of Korea\\
{\tt\small \{jshackist, hjlee0612, rkswlsj13, kateshim\}@kaist.ac.kr}
}

\begin{document}
\maketitle
\begin{abstract}

Existing domain generalization methods for LiDAR semantic segmentation under adverse weather struggle to accurately predict ``things'' categories compared to ``stuff'' categories. 
In typical driving scenes, ``things'' categories can be dynamic 
and associated with higher collision risks, making them crucial for safe navigation and planning. 
Recognizing the importance of ``things'' categories, we identify their performance drop as a serious bottleneck in existing approaches.
We observed that adverse weather induces degradation of semantic-level features and both corruption of local features, leading to a misprediction of ``things'' as ``stuff''.
To mitigate these corruptions, we suggests our method, \textbf{NTN} - segme\textbf{N}t \textbf{T}hings for \textbf{N}o-accident.
To address semantic-level feature corruption, we bind each point feature to its superclass, preventing the misprediction of things classes into visually dissimilar categories.
Additionally, to enhance robustness against local corruption caused by adverse weather, we define each LiDAR beam as a local region and propose a regularization term that aligns the clean data with its corrupted counterpart in feature space.  
NTN achieves state-of-the-art performance with a +2.6 mIoU gain on the SemanticKITTI-to-SemanticSTF benchmark and +7.9 mIoU on the SemanticPOSS-to-SemanticSTF benchmark.
Notably, NTN achieves a +4.8 and +7.9 mIoU improvement on ``things'' classes, respectively, highlighting its effectiveness.

\end{abstract}

\section{Introduction}
\label{sec:intro}

\begin{figure}[ht]
    \centering
    \includegraphics[width=1.0\linewidth]{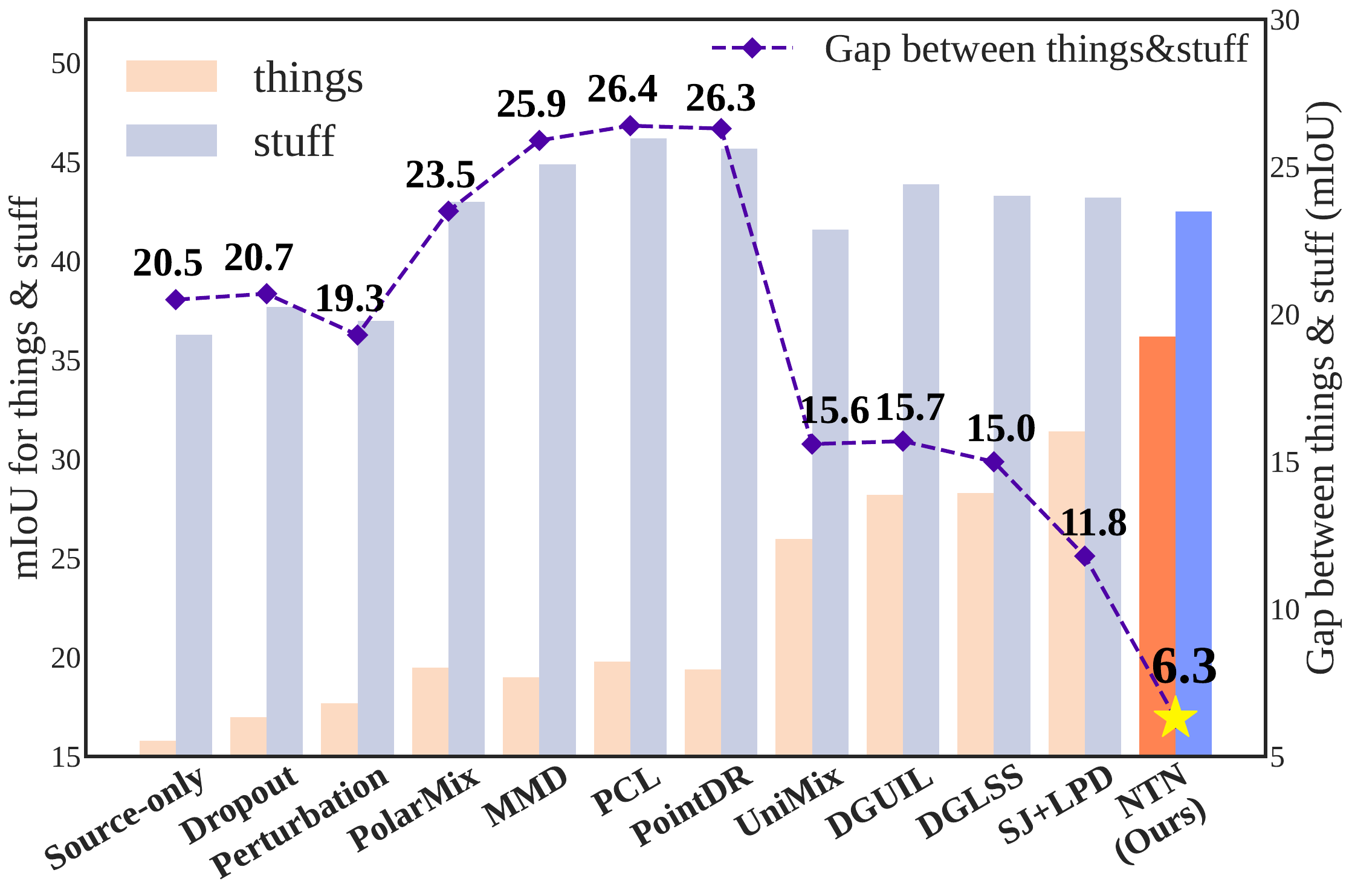}
    \vspace{-1.5em}
    \caption{Previous methods struggle to accurately predict the \textit{things} category in adverse weather. Our approach overcomes this limitation with improved performance for \textit{things} category.}
    \label{fig:motivation}
    \vspace{-1em}
\end{figure}
    % \caption{Differences in point patterns by distance of \textit{car} class. Depending on the distance of the object, the beam that recognizes the object changes, and accordingly, the point pattern changes even for objects of the same category.}

%%%% ====
\begin{figure*}[htb]
    \centering
    \includegraphics[width=\textwidth]{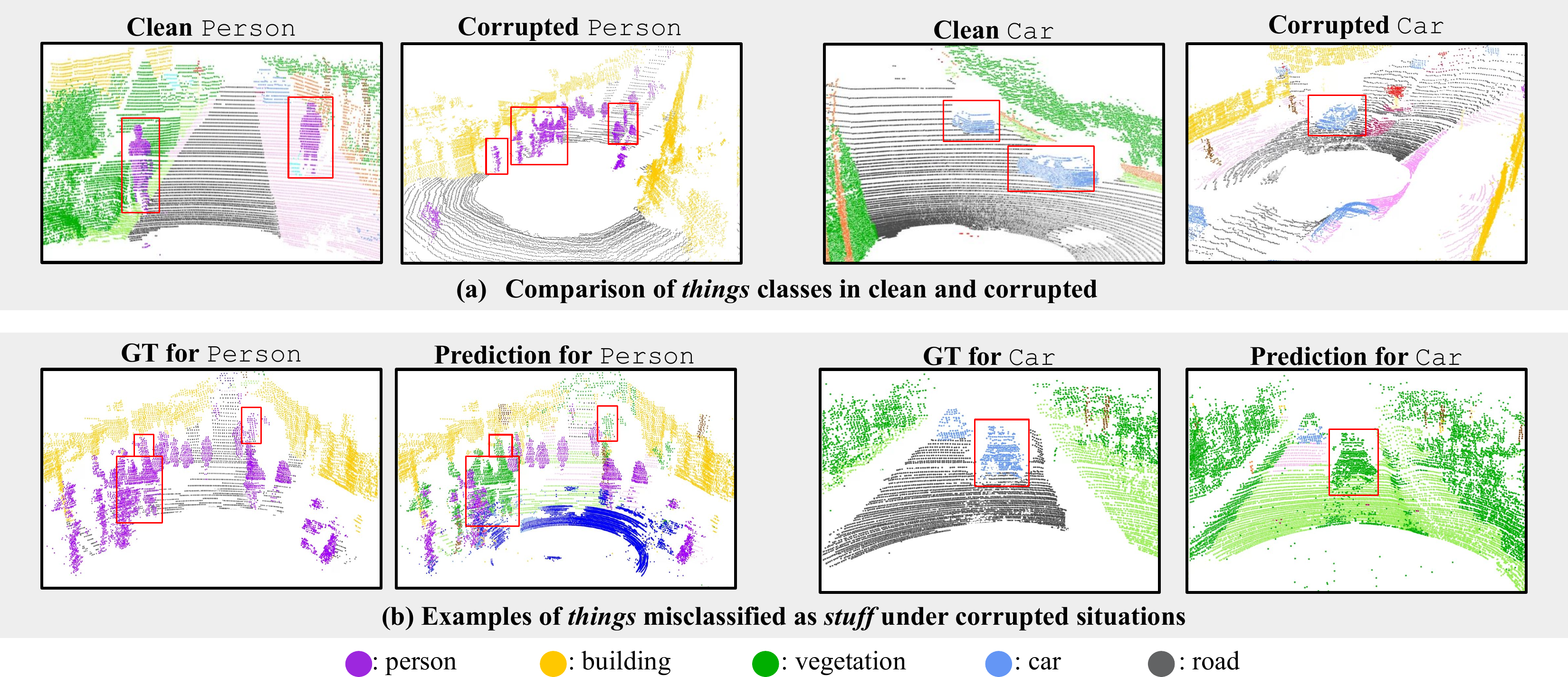}
    \caption{(a) Comparison of \emph{things} classes in SemanticKITTI~\cite{behley2019semantickitti} (clean) and SemanticSTF~\cite{xiao20233d} (corrupted) datasets.
    (b) Examples of \emph{things} classes misclassified as \emph{stuff} due to weather corruption. The prediction results were obtained with SJ+LPD~\cite{park2024rethinking}.
    }
    \label{fig:kitti_stf_compare+misclassified}
\vspace{-0.5em}
\end{figure*}
%%%% ====

% DG4LSSAW
% Method 이름과 제목?

\textbf{L}iDAR \textbf{S}emantic \textbf{S}egmentation (LSS) involves predicting semantic labels directly for each 3D point, an important building block for autonomous driving. However, LSS models exhibit a significant performance drop under adverse weather conditions such as fog, rain, and snow~\cite{xiao20233d, kong2023robo3d, park2024rethinking, zhao2024unimix, he2024domain}. Since safety is the highest priority in autonomous driving, it is essential to develop robust models that can maintain high accuracy regardless of weather conditions.

To mitigate this issue, existing research on improving robustness under adverse weather conditions can be broadly categorized into domain generalization and domain adaptation methods.
Domain adaptation methods~\cite{zhao2024unimix} utilize real weather training data to train LSS models to be robust against weather corruption. 
In contrast, domain generalization methods aim to create models robust to weather corruption only using training data under clean weather. Typically, these studies either improve training mechanisms~\cite{xiao20233d, he2024domain} or generate LiDAR data through weather simulations~\cite{park2024rethinking, hahner2021fog, hahner2022lidar}.
Domain generalization in adverse weather is particularly important, as real weather varies in type and severity and occurs infrequently, making data collection challenging. In this paper, we focus on the domain generalization problem. 

However, existing \textbf{D}omain \textbf{G}eneralization methods for \textbf{L}iDAR \textbf{S}egmentation under \textbf{A}dverse \textbf{W}eather (DGLS-AW) show significant performance variations across semantic classes. In driving scenarios, semantic categories can be divided into \emph{stuff} and \emph{things}. \emph{Stuff} shows continuous distributions and is static such as road and terrain. Meanwhile, \emph{things} consist of distinct and dynamic instances like pedestrians, bicycles, cars, and trucks. As shown in Fig.~\ref{fig:motivation}, the segmentation models generalized for adverse weather~\cite{xiao20233d, kong2023robo3d, park2024rethinking, zhao2024unimix, he2024domain} exhibit significantly lower prediction performance for \emph{things} compared to \emph{stuff}. Even the state-of-the-art model SJ+LPD~\cite{park2024rethinking} shows a significant performance gap, with \emph{things} mIoU averaging 11.8 lower than \emph{stuff}. This performance gap poses serious challenges for autonomous driving. Failing to accurately detect objects like pedestrians or vehicles can lead to critical incidents, as it hinders the ability to react quickly to their movements. On the other hand, missing \emph{stuff} category generally has a lower impact on immediate accident risk. 
\footnote{It is well-known that misclassifying a things category as a stuff category can be fatal, as it fails to account for the potential movement of these objects. It was revealed that an accident that occurred in May 2016 while using a driving assistance system was caused by incorrectly predicting the white part of the trailer as the sky~\cite{kendall2017uncertainties, NHTSA_PE16007_2017}.}
For these reasons, accurately detecting \emph{things} is a critical mission considering driving applications.

Our study focuses on mitigating performance limitations for the \emph{things} classes in DGLS-AW. For that, we first identify key factors contributing to the performance gap and propose a solution to address each challenge: weather-induced corruption confuses the LSS model at both (1) semantic-level and (2) local-level features separately, especially for \emph{things} classes.

First, adverse weather causes severe deformation, particularly affecting \emph{things} classes, making their shapes unrecognizable and hindering the learning of semantic-level features. As shown in Fig.~\ref{fig:kitti_stf_compare+misclassified} (a), \emph{stuff} classes (\texttt{building}, \texttt{vegetation}, and \texttt{road}) are large enough to cover the entire scene, therefore their geometric features remain stable regardless of whether the data is clean or corrupted. In contrast, \emph{things} classes show surprisingly different appearances in clean and corrupted data. 

This difference arises because \emph{things} classes are generally small and dispersed, leaving them vulnerable to missing points that can distort their structure. For example, \texttt{person} class, belonging to the \emph{things} category, have dense and clear shapes under clean weather, as shown on the far left of the Fig.~\ref{fig:kitti_stf_compare+misclassified} (a) and (b). Consequently, under adverse weather conditions, the \texttt{person} and \texttt{car} points are fewer in number and have obscure shapes. We analyze that weather-induced corruption blurs the boundary between \emph{things} and \emph{stuff}, leading to frequent misclassification. Even the state-of-the-art method, SJ+LPD~\cite{park2024rethinking}, often confused between \emph{things} and \emph{stuff} as shown in Fig.~\ref{fig:kitti_stf_compare+misclassified} (b). 

To tackle this, we propose a \textbf{F}eature \textbf{B}inder (FB) that groups semantically similar classes into superclasses. By binding each point feature to its superclass, FB ensures that even in the presence of corruption, \emph{things} classes can learn and retain their semantic information. This approach effectively reduces misclassification between \emph{things} and \emph{stuff}.

Second, the corruption leads to the loss of local-level features, which is especially critical for \emph{things} classes due to their smaller size. Previous research states that adverse weather effects like point drops affect each laser beam independently, causing local feature corruption~\cite{hahner2021fog, hahner2022lidar, park2024rethinking}. Since \emph{things} classes rely heavily on distinctive local features (e.g., point patterns, intensity) for accurate prediction, this corruption is particularly detrimental to them. Missing points remove essential training signals from these regions, preventing effective feature learning for precise predictions. 
% Missing points distort their local-level structures, making it challenging to capture detailed features for precise prediction.

To mitigate this issue, we introduce \textbf{B}eam-wise \textbf{F}eature \textbf{D}istillation (BFD), which aligns the features of corrupted data with those of clean data. By matching corrupted points with their clean counterparts, we provide additional information to the \emph{things} classes. This addition allows the model to robustly predict \emph{things} even under severe corruption, as the dense points from the clean data compensate for the missing information in the corrupted data.

We propose \textbf{NTN} (segme\textbf{N}t \textbf{T}hings for \textbf{N}o-accident), a novel approach that integrates FB and BFD to tackle semantic- and local-level feature corruption in DGLSS-AW.
NTN significantly improves the robustness of LSS models against adverse weather conditions, with particular benefits for safety-critical \emph{things} classes.
This approach achieves state-of-the-art performance on both the SemanticKITTI-to-SemanticSTF and SemanticPOSS-to-SemanticSTF benchmarks, with overall mIoU improvements of +2.6 and +7.9 for each, respectively. Notably, NTN demonstrates a +4.8 and +7.9 boost in \emph{things} classes on these benchmarks, significantly increasing accuracy for highly mobile or structurally complex objects such as \texttt{person}, \texttt{bicycle}, and \texttt{motorcyclist}. These results show NTN’s ability to maintain precise segmentation of risk-sensitive objects in adverse weather scenarios.

\noindent In summary, our contributions are as follows:
\begin{itemize}
    \item[1.] To our knowledge, we are the first to identify a significant performance degradation in safety-critical \emph{things} classes under adverse weather conditions. 
    We found that damage to both semantic- and local-level features caused by weather corruption is a key factor in performance degradation, particularly for \emph{things} classes.
    
    \item[2.] To address these issues, we propose segme\textbf{N}t \textbf{T}hings for \textbf{N}o-accident, which integrates two novel modules; \emph{Feature Binder} and \emph{Beam-wise Feature Distillation}.
    
    \item[3.] Our approach achieves state-of-the-art results on the SemanticKITTI-to-SemanticSTF and SemanticPOSS-to-SemanticSTF benchmarks, with significant gains of+4.8 and +7.9 in \emph{things} classes performance, respectively.
\end{itemize}

\section{Related Works}
\label{sec:relatedworks}

%-------------------------------------------------------------------------
\subsection{LiDAR Semantic Segmentation}
LSS methods can be broadly categorized into (1) point-based, (2) projection-based, and (3) voxel-based approaches based on their input representations.

Point-based methods~\cite{qi2017pointnet, thomas2019kpconv, zhao2021point} process 3D points directly as input. 
PointNet~\cite{qi2017pointnet} employs sequential MLPs for directly merging local features of points into global feature vector.
KPConv~\cite{thomas2019kpconv} defines 3D kernel points and performs direct convolution with input points. 
Building upon this approach, KPConvX~\cite{thomas2024kpconvx} introduces attention mechanisms within the kernel points, thereby enhancing the model capacity without significantly increasing the size of the 3D convolutional network.
The Point-Mixer~\cite{choe2022pointmixer} adapts the MLP-Mixer~\cite{tolstikhin2021mlp} for point cloud processing. 
The Point Transformer~\cite{zhao2021point} leverages a transformer architecture to compute query points within each local neighborhood found by k-nearest neighbors. 
While effective, these methods incur high computational costs due to processing large-scale raw LiDAR data. This is why point-based methods are not suitable for outdoor LSS.

Projection-based methods~\cite{milioto2019rangenet++, ando2023rangevit, kong2023rethinking} project LiDAR points onto a 2D plane, allowing semantic segmentation to leverage architectures originally developed for 2D image tasks.
RangeNet++~\cite{milioto2019rangenet++} was the first to address LSS using range images via spherical projection, adapting 2D convolutional neural networks for this purpose.
RangeViT~\cite{ando2023rangevit} applies a pre-trained ViT model on 2D images, demonstrating effective use of 2D image pre-training for range images. 
RangeFormer~\cite{kong2023rethinking} introduces "RangeAug," which enhances 2D-projected range images by generating multiple inputs to boost model performance. 
Projection-based approaches offer fast inference but are limited by information loss during projection.

Voxel-based methods~\cite{choy20194d, zhou2020cylinder3d, lai2023spherical} divide 3D space into voxel grids for efficient computation. 
MinkowskiNet~\cite{choy20194d} voxelizes LiDAR data into cubic grids, using sparse convolutions. 
SPVCNN~\cite{tang2020searching} builds on MinkowskiNet with an added point-wise MLP. 
Cylinder3D~\cite{zhou2020cylinder3d} uses cylindrical partitions, addressing LiDAR density variations with distance. 
SphereFormer~\cite{lai2023spherical} applies radial windows in voxelization and a transformer to aggregate long-range information. 
Voxel-based methods offer a balance of inference speed and segmentation accuracy.

%-------------------------------------------------------------------------
\subsection{LiDAR under Adverse Weather Conditions}
Recently, LiDAR perception research in adverse weather has gained attention for its practical significance in autonomous driving, focusing on (1) weather simulation and (2) task-agnostic domain generalization methods.

% % Weather simulation methods
Hahner \etal~\cite{hahner2022lidar, hahner2021fog} use physical modeling to create synthetic datasets for specific weather conditions, integrating them into training. 
UniMix~\cite{zhao2024unimix} extends these works for adapting LiDAR segmentation models for adverse weather using these simulations. 
SJ+LPD~\cite{park2024rethinking} deviate from physical modeling, incorporating weather-induced corruption via point dropping and jittering to enhance robustness.

% Task-agnostic domain generalization
Recent task-agnostic approaches~\cite{kong2023robo3d, xiao20233d, he2024domain} employ general machine learning techniques, such as teacher-student frameworks and prototyping, to enhance robustness. 
Kong \etal~\cite{kong2023robo3d} introduced random dropout in the teacher-student architecture to build noise-resistant models. 
PointDR~\cite{xiao20233d} generates a prototype in the clean weather branch and multiplies corrupted branch features by this prototype matrix, utilizing clean data insights. 
DGUIL~\cite{he2024domain} addresses point uncertainty under domain shifts by improving point features across varying distributions in uncertain point clouds.

However, existing weather simulation and domain generalization methods focus on replicating weather conditions and do not address precise recognition of key \emph{things} classes under weather-induced corruption in driving environments. 
Building on SJ+LPD~\cite{park2024rethinking}, our work proposes methods to enhance recognition of \emph{things} classes.

%%%% ====
\begin{figure}[t!]
    \centering
    \includegraphics[width=\linewidth]{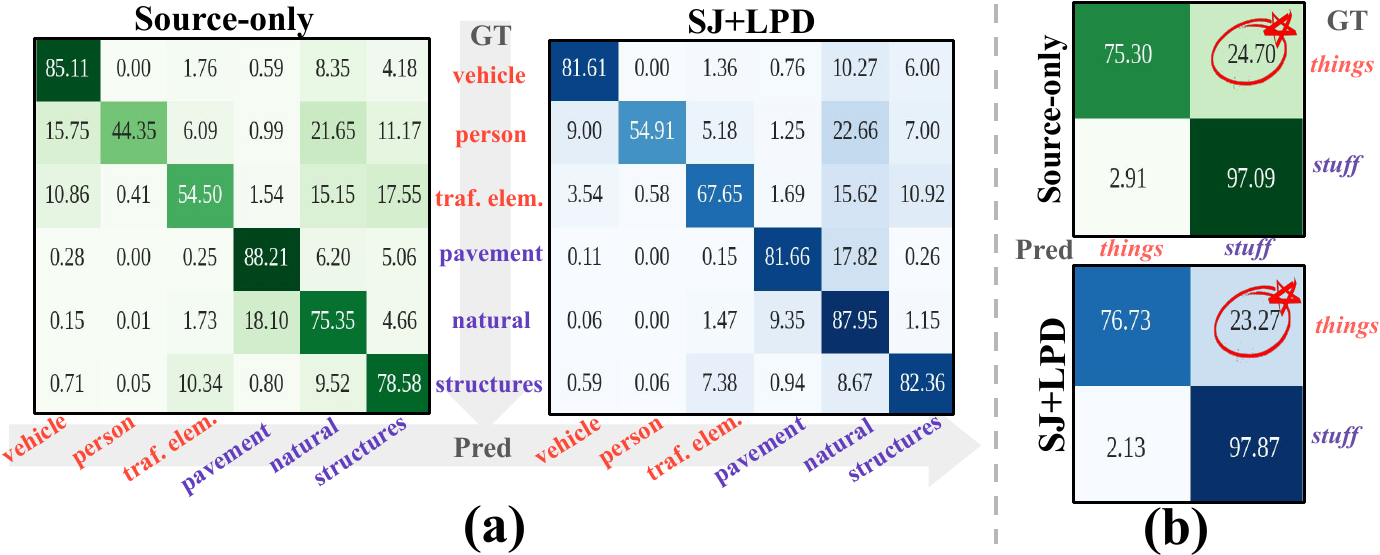}
    \caption{Confusion matrices for (a) superclasses and (b) \textit{things} and \textit{stuff}. It shows a significant performance gap between \textit{things} and \textit{stuff}; \textit{things} are often misclassified as \textit{stuff}, whereas the reverse misclassification is rare.}
    \label{fig:conf}
    %\vspace{-0.8em}
\end{figure}
%%%% ====

\section{Methods}
\label{sec:methods}

\begin{figure*}[th!]
    \centering
    \includegraphics[width=\textwidth]{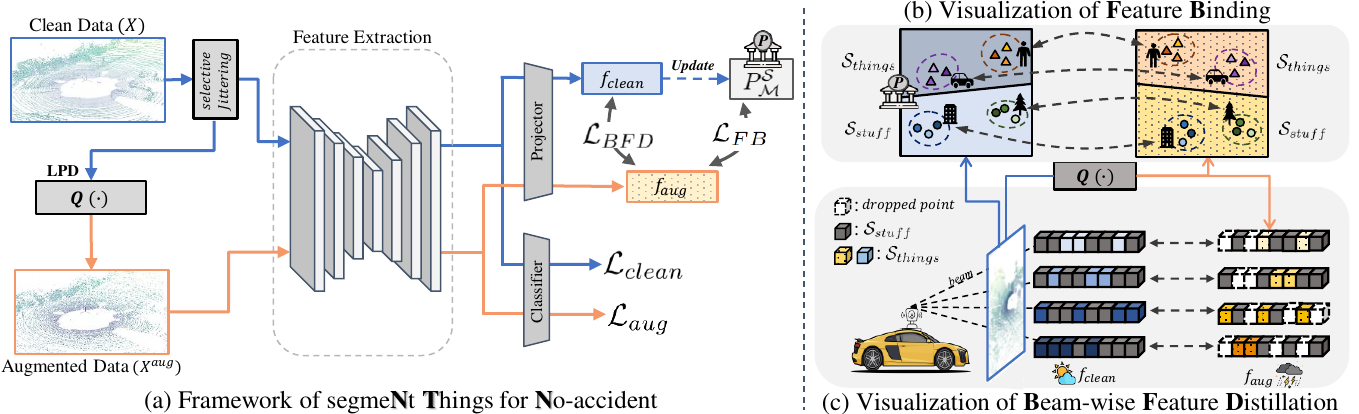}
    \caption{
    (a) Framework of NTN. NTN builds upon SJ+LPD~\cite{park2024rethinking} and enhances the performance of the LSS model on the \emph{things} category through Feature Binding (FB) and Beam-based Feature Distillation (BFD).
    (b) FB stores prototypes obtained from clean data in a memory bank and encourages features derived from augmented data to align with these prototypes. (c) BFD imposes constraints to ensure that features from clean data and those obtained from augmented data match within a manually defined local region, referred to as a beam.}
    \label{fig:overview}
    \vspace{-0.3em}
\end{figure*}
%%--
%% clean data, augmented data 이미지 새로 뽑기
%% Adv. Data -> Augmented Data

%-------------------------------------------------------------------------
\subsection{Motivation}
\label{subsec:method_motivation}
\noindent

In this section, we aim to understand why existing LSS methods in adverse weather underperform on the \emph{things} classes.
To accomplish this, we analyze the impact of weather-induced corruption in adverse weather on 3D points. 
The findings in Sec.~\ref{sec:intro} can be summarized as follows:  
(1) Weather corruption degrades the semantics of \emph{things} classes.  
(2) Local-level feature corruption causes significant missing points and leads to information loss.

% % Confusion Matrix로 보여주기
As shown in Fig.~\ref{fig:conf}, even the current state-of-the-art DGLS-AW method, SJ+LPD~\cite{park2024rethinking}, misclassifies 22.66\% of \texttt{person} predictions as \texttt{natural}, illustrating the tendency for \emph{things} classes' predictions to shift toward \emph{stuff}. This misclassification occurs because the model struggles to maintain clear boundaries between visually distinct classes when the input features are corrupted. 
To address this, we constrain features from the LSS model to remain close within visually similar classes. This approach preserves semantic alignment and mitigates shifts between \emph{things} and \emph{stuff}, even under corruption-induced semantic degradation.

Also, as shown in Fig.~\ref{fig:kitti_stf_compare+misclassified} (a), significant missing points cause a loss of learning signals during training. This loss is particularly harmful to \emph{things} classes.
It is because \emph{things} classes, with fewer points and smaller sizes, heavily rely on these learning signals.
Specifically, in SemanticKITTI~\cite{behley2019semantickitti}, \emph{things} classes account for only 12.3\% of the total points.
The limited points in the \emph{things} classes become even scarcer when missing, resulting the poor performance in the \emph{things} category.  

% From these facts, we observe two key points: 
By focusing on these key points, we address the degraded performance of the \emph{things} category in existing DGLS-AW methods. We adopt the existing adverse weather LSS, SJ+LPD~\cite{park2024rethinking} as a baseline, which achieves state-of-the-art performance through specialized weather simulation augmentations.
Importantly, our proposed framework is architecture-independent and applicable to various baseline models beyond SJ+LPD, as it introduces new training schemes.

%%%%%%%%%%%%%%%%%%%%%%%%%%%%%% PREVIOUS METHOD %%%%%%%%%%%%%%%%%%%%%%%%%%%%%%
\subsection{Weather-corrupted LiDAR Data Generation}
\label{subsec:method_sjlpd}
\noindent
Existing LiDAR weather simulation methods~\cite{hahner2021fog, hahner2022lidar, park2024rethinking} simulate corrupted data $X^{aug}$ and use them as additional training data to achieve generalized performance under adverse weather. Hahner \etal~\cite{hahner2022lidar, hahner2021fog} employ physical equations to synthesize weather-specific LiDAR data for training. 
SJ+LPD~\cite{park2024rethinking}, which we set as a baseline method, generates output points $\hat{X}$ by applying selective jittering to the original clean data. Subsequently, $\hat{X}$ is input into the Learnable Point Drop module $Q(\cdot)$ to produce the final augmented data $X^{aug}$.
In subsequent methods, we utilize $\hat{X}$ as clean data (i.e., the original LiDAR points) and $X^{aug}$ as augmented data (i.e., the simulated or augmented counterpart).

\begin{table}[t!]
\centering
\renewcommand{\arraystretch}{1.3} % 줄 간격 조정
\arrayrulecolor{black}
\begin{adjustbox}{max width=\columnwidth}
\begin{tabular}{l|l|l|l}
\specialrule{1.2pt}{0pt}{3pt}
\textbf{Category} & \textbf{Superclass}                  & \textbf{SemanticKITTI}       & \textbf{SemanticPOSS}                                      \\ \hline

\multirow{3}{*}{Things} & Vehicle    & car, bicycle, motorcycle, truck, bus & car, bicycle                \\ \cline{2-4}
                        & Person     & person, bicyclist, motorcyclist       & pedestrian                      \\ \cline{2-4}
                        & Traffic Element & fence, pole, traffic-sign         & -                           \\ \cmidrule(lr{0.5em}){1-4} 
\multirow{3}{*}{Stuff}  & Pavement   & road, parking, sidewalk              & drivable-surface                        \\ \cline{2-4}
                        & Natural    & other-ground, vegetation, terrain    & walkable                     \\ \cline{2-4}
                        & Structure  & building, trunk                      & -                           \\ 
\specialrule{1.2pt}{0pt}{3pt}
\end{tabular}
\end{adjustbox}
\caption{Superclasses of each category and its constituent classes for SemanticKITTI and SemanticPOSS datasets.}
\label{tab:superclass}
\vspace{-0.5em}
\end{table}

%%%%%%%%%%%%%%%%%%%%%%%%%%%%%% METHOD 1 %%%%%%%%%%%%%%%%%%%%%%%%%%%%%%
\subsection{Feature Binding Loss}
\label{subsec:method_spa}

Building on the analysis in Sec.~\ref{subsec:method_motivation}, we first introduce a concept of \textbf{F}eature \textbf{B}inding loss (FB).
The purpose of FB loss is to ``bind'' the output features of LSS model, preventing misprediction of \emph{things} objects as \emph{stuff} when weather corruption severely affects semantic-level features.

To achieve this, we create prototypes for superclasses by grouping visually similar pre-defined classes during LSS training. Tab.~\ref{tab:superclass} summarizes the superclass categories and their corresponding classes. Six superclasses are defined for SemanticKITTI~\cite{behley2019semantickitti}, while four superclasses are defined for SemanticPOSS~\cite{pan2020semanticposs}. Following the method in ~\cite{kim2023single}, we associated each class with its respective superclass. Classes with significantly different shapes, such as \texttt{trunk} and \texttt{vegetation} are assigned to separate superclasses. Additionally, classes previously treated as background, like a motorcyclist, were reassigned to the \texttt{person} class.

We then constrain the output features of the LSS model for augmented data $x^{aug}$ to remain close to these superclass prototypes. 
This approach prevents severe misprediction into visually dissimilar \emph{stuff} classes, even if the semantic features of input points are significantly degraded by weather corruption.

Specifically, we compute the prototype for each class \( c_i \) as the mean feature of its constituent classes and store it in the memory bank \( \mathcal{M} \).
\begin{equation}
    P_{\mathcal{M}}^{c_i} = \frac{1}{|\hat X_{c_i}|} \sum_{\hat x_{c_i} \in \hat X_{c_i}} G(F(\hat x_{c_i})).
\end{equation}

$F$ is the LSS model, $G$ is a projector head with MLP, and $|\hat X_{c_i}|$ denotes the number of points labeled as $c_i$ in the entire class set $\hat X_{\mathcal{C}}$.
These class-wise prototypes are updated using an exponential moving average during each training iteration.
Subsequently, the superclass prototype \( P^{\mathcal{S}_n}_{\mathcal{M}} \) is defined using the class-wise prototypes belonging to each superclass \( \mathcal{S}_n \) as follows:
\begin{equation}
    P^{\mathcal{S}_n}_{\mathcal{M}} = \frac{1}{|\mathcal{S}_n|} \sum_{c_i \in \mathcal{S}_n} P^{c_i}_{\mathcal{M}}.
\end{equation}
where \( |\mathcal{S}_n| \) represents the number of classes in superclass \( \mathcal{S}_n \).
We then fit the feature output from augmented data to each superclass prototype to compute the FB loss as follows:
{\footnotesize
\begin{equation}
    \mathcal{L}_{FB} = \left\| \frac{1}{|X_{c}^{aug}|} \sum_{x_{c}^{aug} \in X_{c}^{aug}} G(F(x_{c}^{aug})) - \frac{1}{|\mathcal{S}|} \sum_{\mathcal{S}_n \in \mathcal{S}} P_{\mathcal{M}}^{\mathcal{S}_n} \right\|^2.
\end{equation}
}
 
This approach enables the LSS model to bind semantic features to their appropriate superclass by creating prototypes of visually similar, pre-defined classes. This method enables the model to prevent misprediction of \emph{things} as \emph{stuff}, even when handling corrupted inputs.

%%%%%%%%%%%%%%%%%%%%%%%%%%%%%% METHOD 2 %%%%%%%%%%%%%%%%%%%%%%%%%%%%%%

%%%% ====
\begin{figure}[tb]
    \centering
    \includegraphics[width=\columnwidth]{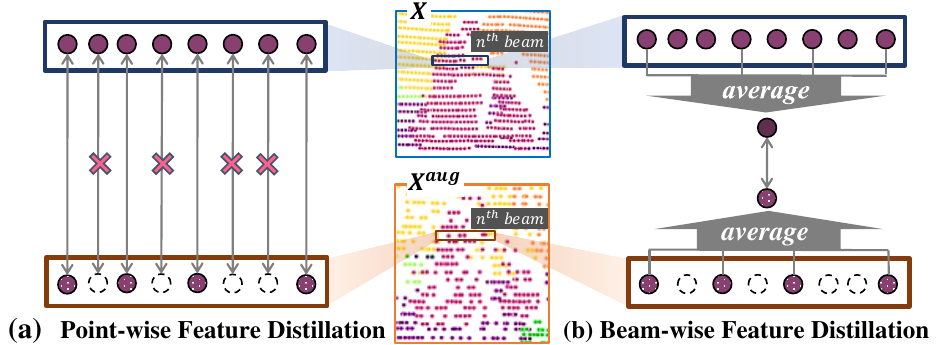}
    \caption{Illustration of (a) Point-wise Feature Distillation and (b) Beam-wise Feature Distillation.}
    \label{fig:beam_avg}
 \vspace{-0.3em}
\end{figure}
%%%% ====

\subsection{Beam-wise Feature Distillation}
\label{subsec:method_bfd}
  \begin{table*}[t!]
\centering
\renewcommand{\arraystretch}{1.1}
\begin{adjustbox}{max width=\textwidth}
\arrayrulecolor{black} % black line
\begin{tabular}{l|c c c c c c c c c c c c c c c c c c c | c} % Method와 mIoU 앞뒤로 수직선 추가
\specialrule{1.3pt}{0pt}{3pt} % 두꺼운 윗줄
\textbf{Method} \rule{0pt}{15pt} & car & bi.cle & mt.cle & truck & oth-v. & pers. & bi.clst & mt.clst & road & parki. & sidew. & othe.g. & build. & fence & veget. & trunk & terra. & pole & traf. & \textbf{mIoU} \\
\cmidrule(lr{0.5em}){1-1} \cmidrule(lr{0.5em}){2-20} \cmidrule(lr{0.5em}){21-21} % 줄 단위로 수직선 유지
Oracle & 89.4 & 42.1 & 0.0 & 59.9 & 61.2 & 69.6 & 39.0 & 0.0 & 82.2 & 21.5 & 58.2 & 45.6 & 86.1 & 63.6 & 80.2 & 52.0 & 77.6 & 50.1 & 61.7 & 54.7 \\
\cmidrule(lr{0.5em}){1-1} \cmidrule(lr{0.5em}){2-20} \cmidrule(lr{0.5em}){21-21}
\arrayrulecolor{black} 
Source-only & 55.9 & 0.0 & 0.2 & 1.9 & 10.9 & 10.3 & 6.0 & 0.0 & 61.2 & 10.9 & 32.0 & 0.0 & 67.9 & 41.6 & 49.8 & 27.9 & 40.8 & 29.6 & 17.5 & 24.4 \\
Dropout~\cite{dropout} & 62.1 & 0.0 & 15.5 & 3.0 & 11.5 & 5.4 & 2.0 & 0.0 & 58.4 & 12.8 & 26.7 & 1.1 & 72.1 & 43.6 & 52.9 & 34.2 & 43.5 & 28.4 & 15.5 & 25.7 \\
Perturbation & 74.4 & 0.0 & 0.0 & 23.3 & 0.6 & 19.7 & 0.0 & 0.0 & 60.3 & 10.8 & 33.9 & 0.7 & 72.0 & 45.2 & 58.7 & 17.5 & 42.4 & 22.1 & 9.7 & 25.9 \\
PolarMix~\cite{xiao2022polarmix} & 57.8 & 1.8 & 3.6 & 3.7 & 26.5 & 3.7 & 26.5 & 0.0 & 65.7 & 2.9 & 35.9 & 48.7 & 71.0 & 48.7 & 53.8 & 20.5 & 45.4 & 25.9 & 15.8 & 26.5 \\
MMD~\cite{li2018domain} & 63.6 & 0.0 & 2.6 & 17.4 & 11.4 & 28.1 & 0.0 & 0.0 & 67.0 & 14.1 & 37.6 & 41.2 & 67.1 & 41.2 & 57.1 & 27.4 & 47.9 & 28.2 & 16.2 & 26.9 \\
PCL~\cite{yao2022pcl} & 65.9 & 0.0 & 0.4 & 17.3 & 8.4 & 8.4 & 8.4 & 8.4 & 59.6 & 12.0 & 35.3 & 63.1 & 74.0 & 47.5 & 60.7 & 15.8 & 48.9 & 26.1 & 27.5 & 26.4 \\
PointDR~\cite{xiao20233d} & 67.3 & 0.0 & 4.5 & 19.9 & 18.8 & 2.7 & 0.0 & 0.0 & 62.6 & 12.9 & 38.6 & 43.8 & 73.3 & 43.8 & 56.4 & 32.2 & 45.7 & 28.7 & 27.4 & 28.6 \\
DGLSS~\cite{kim2023single} & 72.6 & 0.1 & 11.7 & 29.4 & 13.7 & 48.3 & 0.5 & 21.2 & 65.0 & 20.2 & 38.3 & 3.8 & 78.9 & 51.8 & 57.0 & 36.4 & 47.0 & 26.9 & 34.9 & 34.6 \\
UniMix~\cite{zhao2024unimix} & 82.7 & 6.6 & 8.6 & 4.5 & 19.9 & 35.5 & 15.1 & 15.5 & 55.8 & 10.2 & 36.5 & 40.1 & 72.8 & 40.1 & 49.1 & 33.4 & 34.9 & 23.5 & 33.5 & 31.4 \\
DGUIL~\cite{he2024domain} &  77.9 & 10.6 & 19.1 & 26.0 & 9.7 & 46.3 & 6.0 & 9.3 & 69.1 & 18.0 & 38.6 & 9.4 & 73.3 & 51.2 & 60.8 & 30.9 & 50.8 & 31.8 & 22.3 & 35.5 \\
\arrayrulecolor{lightgray} 
\cmidrule(lr{1.em}){1-1} \cmidrule(lr{1.em}){2-20} \cmidrule(lr{1.em}){21-21}
SJ+LPD~\cite{park2024rethinking} & 83.1 & 1.2 & 17.2 & 30.5 & 18.4 & 47.5 & 10.7 & 18.8 & 64.0 & 15.9 & 38.7 & 4.6 & 77.4 & 50.8 & 59.7 & 37.2 & 47.7 & 31.1 & 35.8 & 36.3 \\
\rowcolor{gray!10}
\textbf{NTN (Ours)} &  83.3 & 3.7 & 31.3 & 36.2 & 18.2 & 53.3 & 6.8 & 55.9 & 67.2 & 18.1 & 37.2 & 5.4 & 72.1 & 41.8 & 58.0 & 36.0 & 46.0 & 28.2 & 39.8 & 38.9 \\[-0.3em]
\rowcolor{gray!10}
\textbf{\(\uparrow\) to SJ+LPD} & 
{\footnotesize \textcolor{red}{(+0.2)}} & 
{\footnotesize \textcolor{red}{(+2.5)}} & 
{\footnotesize \textcolor{red}{(+14.1)}} & 
{\footnotesize \textcolor{red}{(+5.7)}} & 
{\footnotesize \textcolor{red}{(-0.2)}} & 
{\footnotesize \textcolor{red}{(+5.8)}} & 
{\footnotesize \textcolor{red}{(-3.9)}} & 
{\footnotesize \textcolor{red}{(+37.1)}} & 
{\footnotesize \textcolor[rgb]{0,0.5,0}{(+3.2)}} & 
{\footnotesize \textcolor[rgb]{0,0.5,0}{(+2.2)}} & 
{\footnotesize \textcolor[rgb]{0,0.5,0}{(-1.5)}} & 
{\footnotesize \textcolor[rgb]{0,0.5,0}{(+0.8)}} & 
{\footnotesize \textcolor[rgb]{0,0.5,0}{(-5.3)}} & 
{\footnotesize \textcolor{red}{(-9.0)}} & 
{\footnotesize \textcolor[rgb]{0,0.5,0}{(-1.7)}} & 
{\footnotesize \textcolor[rgb]{0,0.5,0}{(-1.2)}} & 
{\footnotesize \textcolor[rgb]{0,0.5,0}{(-1.7)}} & 
{\footnotesize \textcolor{red}{(+2.9)}} & 
{\footnotesize \textcolor{red}{(+4.0)}} & 
\textcolor{red}{\textbf{(+2.6)}} \\
\arrayrulecolor{black}
\specialrule{1.3pt}{3pt}{0pt} % 두꺼운 아랫줄
\end{tabular}
\end{adjustbox}
\caption{Comparison of methods on the SemanticKITTI-to-SemanticSTF benchmark. Performance improvements of our method over the current state-of-the-art model, SJ+LPD, are shown in the last row, with red text indicating \emph{things} classes increments and green text for \emph{stuff}.}
\label{tab:main_exp_kitti2stf}
\vspace{-0.2em}
\end{table*}

\noindent
 
This section presents a method to generate additional learning signals for \emph{things} classes, which are particularly vulnerable to weather corruption due to their limited representation in the dataset. Weather corruption primarily affects localized regions, causing significant point loss and disrupting learning signals. Without sufficient learning signals, the model struggles to capture the necessary features for these classes, leading to degraded performance. To address this issue, we design an additional loss that also operates within localized areas, compensating for the loss of learning signals. 

To define the local regions, we slice the LiDAR scan into each LiDAR beam. A beam corresponds to the output response of the laser beam that the LiDAR sensor fires at regular pitch angle intervals. 
Each beam scans the \emph{things} object into upper and lower local regions. Using these beams, we define a loss ensuring that the average of the clean and augmented data is identical for each beam, exclusively for the \emph{things} classes.

Specifically, we first apply off-the-shelf beam separation algorithms~\cite{hahner2021fog, hahner2022lidar, kong2023robo3d} for LiDAR data to identify the LiDAR beam source of the points corresponding to each voxel. Then, for each \emph{things} classes and each LiDAR beam, we apply an MSE loss between the feature of clean data and that of augmented data.

{\footnotesize
    \begin{equation}
    \mathcal{F}_{mean}^{c, b} = \frac{1}{N_{c, b}} \sum_{i=1}^{N_{c, b}} F(\hat x_i), \quad 
    \mathcal{F}_{mean}^{'c, b} = \frac{1}{N_{c, b}^{'}} \sum_{i=1}^{N_{c, b}^{'}} F(x_i^{aug}).
    \end{equation}
}
{\footnotesize
    \begin{equation}
    \mathcal{L}_{BFD} = \sum_{c \in \mathcal{S}_{\text{things}}} \sum_{b \in \mathcal{B}} \left\| \mathcal{F}_{mean}^{c, b} - \mathcal{F}_{mean}^{'c, b} \right\|^2.
    \end{equation}
}
 
\(\mathcal{S}_{\text{things}}\) denote the \emph{things} classes, and \(c\) a specific class. Let \(\mathcal{B}\) represent all LiDAR beams, and \(b\) a specific beam. \(N_{c, b}\) is the number of points for class \(c\) and beam \(b\).
\(\mathcal{L}_{BFD}\) minimizes the difference between the average features of clean data, \(\mathcal{F}_{mean}^{c, b}\), and that of augmented data, \(\mathcal{F}_{mean}^{'c, b}\), for each beam, specifically for the \emph{things} classes, as shown in Fig.~\ref{fig:beam_avg} (b).

Our proposed beam-wise feature matching focuses on local regions, enabling learning from all points in $\hat{X}$ and leveraging contextual information around each beam unlike point-wise feature distillation in Fig.~\ref{fig:beam_avg} (a). Points-wise feature distillation matches individual point features for distillation without averaging. In this case, no distillation loss can be computed for points dropped from $\hat{X}$ when obtaining $X^{aug}$. Furthermore, surrounding points contribute no learning information, limiting the clean data branch to learning only from a subset of points in the augmented data branch. In contrast, our beam-wise feature matching overcomes these limitations by effectively handling local feature corruptions in \emph{things} category objects, both before and after weather-induced corruption. Therefore, BFD is a effective solution for handling local feature corruption in \emph{things} category objects, both before and after weather-induced corruption.

\subsection{Overall Pipeline}
\label{subsec:overall_pipeline}
 
The overall training pipeline is illustrated in Fig.~\ref{fig:overview}.
The process for obtaining clean and augmented data follows SJ+LPD~\cite{park2024rethinking}.
In SJ+LPD, Selective Jittering (SJ) is applied to obtain clean data $\hat x_i$. The loss and entropy from this data are used as the state for the Learnable Point Drop (LPD).
LPD then drops points, and the LSS model outputs features for the augmented data $x_i^{aug}$.
Thus, we obtain the output feature of clean data \(F(\hat x_i)\) and that of augmented data \(F(x_i^{aug})\).
These features are used to compute the FB loss and BFD loss, as described earlier.
Additionally, we include the cross-entropy loss from both clean and augmented data.
The overall loss is:

\begin{equation}
\begin{split}
\mathcal{L}_{total} = &\ \mathcal{L}_{clean} + \lambda_{aug} \mathcal{L}_{aug} \\
& + \lambda_{FB} \mathcal{L}_{FB} + \lambda_{BFD} \mathcal{L}_{BFD}.
\end{split}
\end{equation}

where $\lambda_{\cdot}$ means the weight for each loss.

\section{Experiments}
\label{sec:experiments}

%-------------------------------------------------------------------------
\subsection{Experimental Setup}

\noindent
\textbf{Dataset.}
We utilize three datasets for autonomous driving to train and evaluate the generalization performance of the LSS model under adverse weather conditions. 
SemanticKITTI dataset uses a 64-beam LiDAR and provides 19 class labels. Following the official sequence split, we use scenes 00-07 and 09-10 for training, comprising 19,130 frames. The SemanticPOSS dataset uses a 40-beam LiDAR and focuses on campus scenes with a high density of cars, riders, and pedestrians. This dataset provides 14 class labels, and we use scenes 00-02 and 04-05 for training, comprising 2,488 frames. SemanticSTF dataset is collected under real-world adverse weather conditions, including rain, fog, and snow. This dataset follows the same 19 class labels used in the SemanticKITTI benchmark. We validate our model on the SemanticSTF \textit{validation set} to evaluate robustness in adverse weather. 

\noindent
\textbf{Class mappings.} We apply different class mappings depending on the generalization scenario. For SemanticKITTI $\rightarrow$ {SemanticSTF}, we use all 19 classes since they share identical classes. For SemanticPOSS $\rightarrow$ {SemanticSTF}, we follow the training and evaluation settings of DGLSS~\cite{kim2023single}, using only 5 classes: \{\texttt{car}, \texttt{bicycle}, \texttt{pedestrian}, \texttt{road}, \texttt{walkable}\}.

\noindent
\textbf{Evaluation Metrics.} We adopt Intersection over Union (IoU) as the evaluation metric for each segmentation class and calculate the mean IoU (mIoU) across all classes.

\noindent
\textbf{Implementation Details.}
% - backbone: MinkowskiNet18~\cite{choy20194d}
% - four A6000 GPUS
% - epoch: 30 / batch 2
% - optimizer: SGD  / learning rate: 0.24
% - loss weight 
% \lambda_{aug.}: 0.5
% \lambda_{GPA}: 0.5
% \lambda_{BFD}: 0.1
We adopted MinkowskiNet-18 architectures~\cite{choy20194d} as our baseline models. 
The training was conducted using a stochastic gradient descent (SGD) optimizer with a learning rate of 0.24 and a weight decay of 0.0001. 
The configurations for SJ+LPD are the same as the previous work~\cite{park2024rethinking}.
All experiments were executed on four NVIDIA A6000 GPUs for 30 epochs with a batch size of 2. 
The total runtime for each experiment ranged from 6 to 8 hours. 
The loss function weights were set to $\lambda_{\text{aug}} = 0.5$, $\lambda_{\text{SPA}} = 0.5$, and $\lambda_{\text{BFD}} = 0.1$.

\begin{table}[t]
\centering
\renewcommand{\arraystretch}{1.03} % 줄 간격 조정
\arrayrulecolor{black}
\begin{adjustbox}{max width=\columnwidth}
\begin{tabular}{lcccccc}
\specialrule{1.2pt}{0pt}{3pt}
\textbf{Method} & Car & Bicycle & Person & Road & Walkable & \textbf{mIoU} \\
\cmidrule(lr{0.5em}){1-1} \cmidrule(lr{0.5em}){2-6} \cmidrule(lr{0.5em}){7-7}
Source-only~\cite{xiao20233d} & 46.6 & 3.1 & 54.2 & 47.0 & 37.2 & 37.6 \\
PointDR~\cite{xiao20233d} & 35.0 & 3.1 & 59.7 & \textbf{60.7} & 14.9 & 34.7 \\
DGLSS~\cite{kim2023single} & 60.0 & 3.1 & 38.6 & 54.0 & 40.1 & 39.2 \\
SJ+LPD~\cite{park2024rethinking} & \textbf{67.0} & 0.9 & 45.9 & 43.3 & 34.7 & 38.3 \\
\rowcolor{gray!10}
\textbf{NTN (Ours)} & 65.5 & \textbf{5.8} & \textbf{61.4} & 54.6 & \textbf{43.6} & \textbf{46.2} \\ [-0.3em]
\rowcolor{gray!10}
\textbf{\(\uparrow\) to SJ+LPD} & 
{\footnotesize \textcolor{red}{(-1.5)}} & 
{\footnotesize \textcolor{red}{(+4.9)}} & 
{\footnotesize \textcolor{red}{(+15.5)}} & 
{\footnotesize \textcolor[rgb]{0,0.5,0}{(+11.3)}} & 
{\footnotesize \textcolor[rgb]{0,0.5,0}{(+8.9)}} & 
\textcolor{red}{\textbf{(+7.9)}} \\
\specialrule{1.2pt}{0pt}{3pt}
\end{tabular}
\end{adjustbox}
\vspace{-1em}
\caption{Comparison of methods on the SemanticPOSS-to-SemanticSTF benchmark.}
\label{tab:main_exp_poss2stf}
\vspace{-1.5em}
\end{table}

\subsection{Main Results}
\noindent\textbf{SemanticKITTI to SemanticSTF.} As shown in Table~\ref{tab:main_exp_kitti2stf}, our method achieves a state-of-the-art performance with a \textbf{\textbf{+2.6}} mIoU improvement over the prior leading model, SJ+LPD~\cite{park2024rethinking}. Additionally, our approach shows a significant boost of +14.5 mIoU over the source-only model, the original MinkUnet. Furthermore, our method achieves notable improvements across most \emph{things} classes, yielding gains of \textbf{+2.5, +14.1, +5.7, +5.8, +37.1, +2.9,} and \textbf{+4.0} mIoU in \texttt{bicycle, motorcycle, truck, person, motorcyclist, pole,} and \texttt{traffic sign} classes, respectively.
These results provide strong evidence that our method significantly enhances performance specifically for \emph{things} classes, as intended. Notably, previous methods struggled to predict tail classes with small-sized, such as \texttt{motorcycle} and \texttt{motorcyclist}, accurately. It indicates that our method offers precise and reliable signals for identifying \emph{things} classes under adverse conditions.

\begin{table}[t!]
\centering
\renewcommand{\arraystretch}{1.05} % 줄 간격 조정
\arrayrulecolor{black}
\begin{adjustbox}{max width=0.88\columnwidth}
\begin{tabular}{c|c|c|lll}
\specialrule{1.2pt}{0pt}{3pt}
\multirow{2}{*}{\textbf{Prototype}} & \multicolumn{2}{c|}{\textbf{Feature Distillation}} & \multicolumn{3}{c}{\textbf{Performance}} \\
 & Matching & Classes & \textbf{Things} & \textbf{Stuff} & \textbf{mIoU} \\
\cmidrule(lr{0.5em}){1-1} \cmidrule(lr{0.5em}){2-3} \cmidrule(lr{0.5em}){4-6}
  \xmark & \xmark & \xmark & 31.4 & 43.2 & 36.3 \\ % baseline (SJ+LPD)
  \arrayrulecolor{lightgray}
 \cmidrule(lr{0.5em}){1-6}
 Class-wise & \xmark & \xmark & 31.6 & 42.2 & 36.1 \\
 \rowcolor{gray!10}
 Superclass &  \xmark & \xmark & 34.9 \textcolor{red}{\scriptsize (\(\uparrow\)3.5)} & 42.5 & 38.1 \textcolor{red}{\scriptsize (\(\uparrow\)1.8)} \\
 Coarse & \xmark & \xmark & 35.6 & 41.0 & 37.9 \\
 \cmidrule(lr{0.5em}){1-6}
 \xmark & beam-wise & all & 34.0 & 43.3 & 37.9 \\
 \xmark & point-wise & things & 33.1 & 41.7 & 36.7 \\
  \rowcolor{gray!10}
  \xmark & beam-wise & things & 34.4 \textcolor{red}{\scriptsize (\(\uparrow\)3.0)} & 44.0 & 38.4 \textcolor{red}{\scriptsize (\(\uparrow\)2.1)} \\
  \cmidrule(lr{0.5em}){1-6}
  \rowcolor{gray!10}
 Superclass & beam-wise & things & \textbf{36.2} \textcolor{red}{\scriptsize (\(\uparrow\)4.8)} & 42.5 & \textbf{38.9} \textcolor{red}{\scriptsize (\(\uparrow\)2.6)} \\
 \arrayrulecolor{black}
\specialrule{1.2pt}{0pt}{3pt}
\end{tabular}
\end{adjustbox}
\caption{Ablation study for different prototype and feature distillation settings. Coarse refers to the use of two prototypes divided into \emph{things} and \emph{stuff} for feature binding}
\label{tab:ablation}
\vspace{-0.8em}
\end{table}

\noindent\textbf{SemanticPOSS to SemanticSTF.}
As shown in Table~\ref{tab:main_exp_poss2stf}, our method establishes a new state-of-the-art on the SemanticPOSS-to-SemanticSTF benchmark, surpassing the performance of the previous model, SJ+LPD~\cite{park2024rethinking}, by an impressive \textbf{+7.9} mIoU. Importantly, in the \texttt{person} category, our method achieves a \textbf{+15.5} mIoU gain over SJ+LPD. These results underscore the effectiveness of our approach in providing highly accurate segmentation signals for \emph{things} classes under challenging adverse conditions.
The SemanticPOSS dataset contains a higher proportion of \emph{things} classes at 28.9\%, compared to SemanticSTF, and is known for its diversity. Consequently, the model has already learned relatively rich features, resulting in less performance improvement for \emph{things} classes compared to SemanticKITTI-to-SemanticSTF benchmark.

\begin{figure*}[t!]
    \centering
    \includegraphics[width=\textwidth]{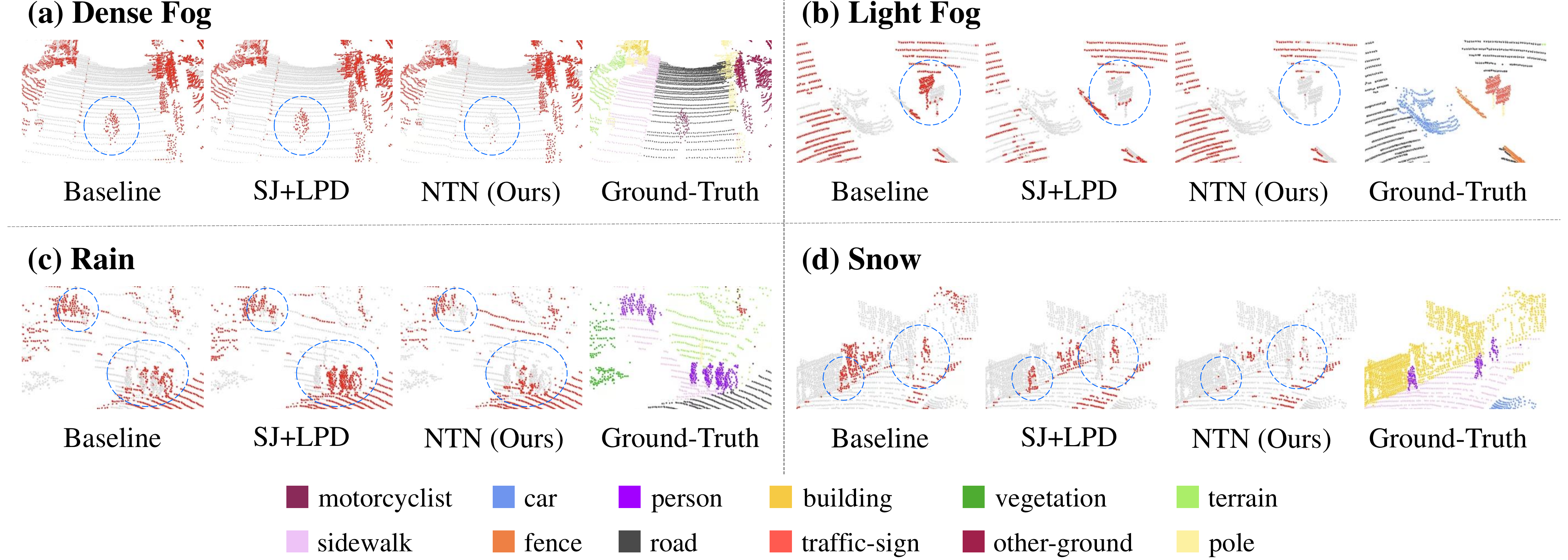}
    \caption{Qualitative results of our method on \textit{validation set} of SemanticSTF. All models are trained on the \textit{train set} of SemanticKITTI. Gray points indicate correct predictions, red points highlight errors, and ground-truth is shown with color-coded labels. Dashed circles highlight the predictions of \textit{things} classes. Best viewed in color.}
    \label{fig:qualitative}
    %\vspace{-0.5em}
\end{figure*}

\subsection{Ablataion Study}
All ablation studies are conducted on the SemanticKITTI-to-SemanticSTF benchmark. 

\noindent\textbf{Feature Binding Loss.}
We conduct an ablation study to evaluate the impact of various prototype strategies of Feature Binding. We compare three approaches: assigning prototypes to each class individually, grouping semantically similar classes into superclasses, and broadly categorizing classes into coarse prototypes as \emph{stuff} or \emph{things}.
As shown in Table~\ref{tab:ablation}, using superclass prototypes improves \emph{things} classes performance by \textbf{+3.3} mIoU compared to the class-wise method. 
This improvement indicates that our FB method effectively bounds \emph{things} classes, which are prone to be misclassified as \emph{stuff}, within their respective superclasses.
The coarse prototype approach performs well for \emph{things} classes but reduces \emph{stuff} class performance due to the use of a single prototype for numerous samples. In contrast, the superclass prototype strategy achieves balanced results across both \emph{things} and \emph{stuff}, providing a \textbf{+3.5} mIoU boost in \emph{things} classes. 
This shows that creating prototypes by grouping classes with high visual similarity into superclasses is more effective for feature binding.
Although the coarse prototype method enhances \emph{things} classes, it causes a 1.5 mIoU decrease in \emph{stuff} and a 0.2 mIoU drop in overall mIoU compared to the superclass prototype approach. This result shows that combining classes with low visual similarity into a single prototype provides suboptimal gains. Besides, improved prediction of \emph{things} in driving scenes is beneficial, but poor prediction of \emph{stuff} can disrupt long-term planning. Thus, the coarse prototype approach cannot be considered to produce ideal results.

\noindent\textbf{Beam-wise Feature Distillation.}
We compare performance based on how feature distillation is applied and to which classes. Table~\ref{tab:ablation} shows that limiting BFD to \emph{things} classes provides better performance across both \emph{things} and \emph{stuff} compared to applying BFD to all classes. This likely occurs because \emph{stuff} classes, with more points in the pitch direction, create class imbalance when BFD is applied. Additionally, point-wise feature matching in row 6 significantly reduces \emph{things} classes performance compared to beam-wise distillation, as it fails to capture learning signals from dropped points, as noted in Sec.~\ref{subsec:method_bfd}.

\noindent\textbf{Overall Comparison.}
Applying superclass prototypes with beam-wise feature distillation only to \emph{things} classes improves mIoU by 2.6 mIoU over the previous SoTA model, with a remarkable 4.8 mIoU increase in \emph{things} classes performance.
% 최종적으로 Superclass prototype과 things class에 대해서만 beam-wise feature distillation를 주었을 때 이전 SoTA 모델대비 mIoU가 2.6 mIoU 올라 가장 좋은 성능을 보이고 있으며, 특히 things class의 성능이 4.8 mIoU로 크게 오른 것을 확인할 수 있다.

\subsection{Qualitative Results}
Fig.~\ref{fig:qualitative} shows qualitative results of our method and baselines on the SemanticSTF validation set, using SemanticKITTI as the source domain. Correct predictions are marked in gray, errors are in red, and ground-truth classes are color-coded. Notably, our method outperforms SJLPD in segmenting \emph{things} category, underscoring its intended gains for these critical classes. Specifically, in the dense fog scenario (a), our method accurately segments the \texttt{motorcyclist} class, while the original MinkowskiNet and SJ+LPD struggle to recognize it. This aligns with our quantitative result, with a 37.1 IoU gain for the \texttt{motorcyclist} class. Additionally, in rain and snow scenarios (c-d), our model excels in segmenting the \texttt{person} class, a critical class for autonomous driving due to its direct impact on safety.

\section{Conclusion}
\label{sec:conclusion}
In this study, we introduced our method, NTN, to address the limitations of existing DGLS-AW, particularly their poor performance on \emph{things} classes. We identified two primary causes of this issue: (1) corruption of semantic-level features in \emph{things} objects and (2) their vulnerability to local feature corruption caused by adverse weather conditions.
To alleviate semantic-level feature corruption, we proposed a feature-binding approach that leverages superclass prototypes. To address the vulnerability of \emph{things} classes to corruptions on local features, we introduced a beam-wise feature distillation loss that utilizes clean-weather beam-level features. Our methods establish a state-of-the-art on the SemanticKITTI-to-SemanticSTF and SemanticPOSS-to-SemanticSTF benchmarks, significantly boosting \emph{things} classes performance.

% [Acknowledgements]
\noindent\textbf{Acknowledgements}. 
This research was supported by the Basic Science Research Program through the National Research Foundation of Korea (NRF) funded by the MSIP (RS-2025-00520207, RS-2023-00219019), KEIT grant funded by the Korean government (MOTIE) (No. 2022-0-00680, No. 2022-0-01045), the IITP grant funded by the Korean government (MSIT) (No. RS-2024-00457882, National AI Research Lab Project, No. 2021-0-02068 Artificial Intelligence Innovation Hub, RS-2019-II190075 Artificial Intelligence Graduate School Program (KAIST)), Samsung Electronics Co., Ltd (IO230508-06190-01) and SAMSUNG Research, Samsung Electronics Co., Ltd.
% WARNING: do not forget to delete the supplementary pages from your submission 
% % 대충 생각하고 있는 내용
% \url{https://docs.google.com/document/d/1xIGmqrEL1_BeBdNvhRy79TUtUBQAUWPg61w0XIw60EE/edit?usp=sharing}

\clearpage
\setcounter{page}{1}
\maketitlesupplementary

\renewcommand{\thefigure}{S\arabic{figure}}
\renewcommand{\thetable}{S\arabic{table}}
\renewcommand{\thesection}{S\arabic{section}}
\renewcommand{\thesubsection}{S\arabic{section}.\arabic{subsection}}

\noindent In this supplementary material, we provide additional details and results not included in the main paper due to space constraints. The content is organized in the following order:

\vspace{0.5em}
\begin{itemize}
    \item Sec.~\ref{sec:rationale}. summarizes the rationale of our methods, Feature Binding and Beam-wise Feature Distillation.
    
    \item Sec.~\ref{sec:weather}. provides results across various weather conditions.

    \item Sec.~\ref{sec:quali_poss}. offers qualitative results on the SemanticPOSS-to-SemanticSTF benchmark.
    
    \item Sec.~\ref{sec:synlidar}. provides experimental results of the SynLiDAR-to-SemanticSTF benchmark.

    \item Sec.~\ref{sec:analysis}. provides discussions about the effectiveness and limitations of our methods.

    % \item Sec.~\ref{sec:analysis_bfd} provides more analysis about Beam-wise Feature Distillation.
    
    \item Sec.~\ref{sec:bicyclist_miou} analyzes the mIoU drop for the \texttt{bicyclist} and \texttt{fence} class.

    \item Sec.~\ref{sec:car_poss2stf}. provides discussions about the performance degradation of \texttt{car} class on SemanticPOSS-to-SemanticSTF benchmark.
    
    \item Sec.~\ref{sec:related_works_more}. provides related works on subclass- or prototype- based methods.
    
    \item Sec.~\ref{sec:weather_simulation_more}. provides additional experiments on different weather simulations.
    
    \item Sec.~\ref{sec:experiments_more}. provides additional experiments on various datasets.
    
    \item Sec.~\ref{sec:superclass_more}. provides additional experiments for different superclass criterions.

    % \item Sec.~\ref{sec:non-beam-wise}. contains results of further ablation studies on two different types of feature distillation.

    \item Sec.~\ref{sec:examples}. provides additional examples illustrating the comparison between clean and corrupted data.

    \item Sec.~\ref{sec:failure_case}. provides failure cases of our methods.
\end{itemize}

\vspace{1em}
\section{Rationale}
\label{sec:rationale}

In this section, we provide a detailed explanation of the rationale behind the two proposed methods, as introduced in the main paper.

\vspace{0.5em}

\noindent\textbf{Feature Binding.} As outlined in the main paper, \textbf{F}eature \textbf{B}inding (FB) aims to prevent \emph{things} objects from being mispredicted as \emph{stuff} classes.
\emph{Things} objects are often misclassified as \emph{stuff} due to semantic-level corruption caused by weather perturbations. 
In such conditions, accurately predicting fine-grained classes (\eg \texttt{person}, \texttt{motorcyclist}) with LiDAR Semantic Segmentation models becomes highly challenging. 
FB mitigates this issue by constraining features to visually similar superclasses, reducing \emph{things}-to-\emph{stuff} mispredictions.

\vspace{0.5em}

\noindent\textbf{Beam-wise Feature Distillation.} 
As discussed in the main paper, the goal of \textbf{B}eam-wise \textbf{F}eature \textbf{D}istillation (BFD) is to recover information lost due to missing points.
BFD specifically addresses severe information loss in \emph{things} objects by ensuring features from the augmented branch effectively capture \emph{things} information from the clean branch.
This approach aligns intact point patterns before the drop with collapsed point patterns after the drop, enabling efficient utilization of point pattern information.

\noindent\textbf{Why Divide Corruptions into Semantic and Local Levels?} 
This study categorizes corruptions based on the extent of point loss induced by weather corruption, which indicates potential degradation in data quality.
A significant point loss can severely degrade semantic information. Thus, compensation strategies, such as Feature Binding, must be devised to mitigate this degradation at the semantic level.
Even minimal point corruption can cause information loss. It disproportionately affects small-scale \emph{things} objects. Beam-wise Feature Distillation is introduced to mitigate these corruptions.

\noindent\textbf{Why FB Benefits \emph{things} Classes More?}  
Before applying our methodology, many misclassifications occurred between \emph{things} and \emph{stuff} classes. Thus, providing an additional discriminative signal via Feature Binding (FB) reduces misclassification between \emph{things} and \emph{stuff}. 
Second, we train the model to exploit the common point pattern in \emph{things} classes through FB. This approach helps distinguish \emph{things} from \emph{stuff} because \emph{things} share similar visual patterns, while \emph{stuff} varies widely. 
Furthermore, as shown in Table~\textcolor{red}{4}, classwise prototypes do not improve performance.  
Therefore, our method's effectiveness does not arise from alleviating class imbalance between \emph{things} and \emph{stuff}.

% As discussed in the main paper, the goal of \textbf{B}eam-wise \textbf{F}eature \textbf{D}istillation (BFD) is to recover information lost due to missing points, unrelated to things-to-stuff misclassification.
% % As a result, objects with relatively well-maintained point patterns after applying BFD exhibit more accurate predictions.

%%% ================================================================================
%%% ================================================================================
\vspace{1em}
\section{Results on Specific Weather Conditions}
\label{sec:weather}

SemanticSTF consists of four adverse weather conditions: dense fog, light fog, rain, and snow. We examine the performance under each weather condition. As shown in Table~\ref{tab:weather_performance_kitti}, on the SemanticKITTI-to-SemanticSTF benchmark, we observe that our model produces robust results regardless of the weather condition. Additionally, Table~\ref{tab:weather_performance_poss} shows the results on the SemanticPOSS-to-SemanticSTF benchmark. Our proposed model achieves the highest mIoU in most weather conditions, including significant improvements in rain and snow scenarios. Therefore, our model demonstrates consistent performance in most adverse weather scenarios across both benchmarks.

%%%%%%%%%%%%%%%%%%%%%%%%%% SEMANTICKITTI 2 SEMANTICSTF WEATHER %%%%%%%%%%%%%%%%%%%%%%%%%%
%%%%%%%%%%%%%%%%%%%%%%%%%% SEMANTICKITTI 2 SEMANTICSTF WEATHER %%%%%%%%%%%%%%%%%%%%%%%%%%
\begin{table}[t!]
\centering
\renewcommand{\arraystretch}{1.2} % Adjust row spacing
\begin{adjustbox}{max width=\columnwidth}
\begin{tabular}{lccccc}
\toprule
\textbf{Method} & \textbf{D-fog} & \textbf{L-fog} & \textbf{Rain} & \textbf{Snow} & \textbf{mIoU} \\
\cmidrule(lr{0.5em}){1-1} \cmidrule(lr{0.5em}){2-5} \cmidrule(lr{0.5em}){6-6}
PolarMix~\cite{xiao2022polarmix} & 29.7 & 25.0 & 28.6 & 25.6 & 27.2 \\
PCL~\cite{yao2022pcl} & 28.9 & 27.6 & 30.1 & 24.6 & 27.8 \\
MMD~\cite{li2018domain} & 30.4 & 28.1 & 32.8 & 25.2 & 29.1 \\
PointDR~\cite{xiao20233d} & 31.3 & 29.7 & 31.9 & 26.2 & 29.8 \\
DGLSS~\cite{kim2023single} & 34.2 & 34.8 & \textbf{36.2} & 32.1 & 34.3 \\
UniMix~\cite{zhao2024unimix} & 34.8 & 30.2 & 34.9 & 30.9 & 31.4 \\
DGUIL~\cite{he2024domain} & \textbf{36.3} & 34.5 & 35.5 & \textbf{33.3} & 34.8 \\
SJ+LPD~\cite{park2024rethinking} & 33.9 & \textbf{35.5} & \underline{35.8} & 32.1 & \underline{36.3} \\
\rowcolor{gray!10}
\textbf{NTN (Ours)} & \underline{35.3} & \underline{35.1} & 35.7 & \underline{32.4} & \textbf{38.9} \\
\bottomrule
\end{tabular}
\end{adjustbox}
\caption{Performance comparison of different methods under varying weather conditions on SemanticKITTI-to-SemanticSTF benchmark. \textbf{Bold} indicates the best mIoU and \underline{underlined} indicates the second-best performance.}
\label{tab:weather_performance_kitti}
\end{table}
%%%%%%%%%%%%%%%%%%%%%%%%%% SEMANTICKITTI 2 SEMANTICSTF WEATHER %%%%%%%%%%%%%%%%%%%%%%%%%%
%%%%%%%%%%%%%%%%%%%%%%%%%% SEMANTICKITTI 2 SEMANTICSTF WEATHER %%%%%%%%%%%%%%%%%%%%%%%%%%

%%%%%%%%%%%%%%%%%%%%%%%%%% SEMANTICPOSS 2 SEMANTICSTF WEATHER %%%%%%%%%%%%%%%%%%%%%%%%%%
%%%%%%%%%%%%%%%%%%%%%%%%%% SEMANTICPOSS 2 SEMANTICSTF WEATHER %%%%%%%%%%%%%%%%%%%%%%%%%%
\begin{table}[t!]
\centering
\renewcommand{\arraystretch}{1.2} % Adjust row spacing
\begin{adjustbox}{max width=\columnwidth}
\begin{tabular}{lccccc}
\toprule
\textbf{Method} & \textbf{D-fog} & \textbf{L-fog} & \textbf{Rain} & \textbf{Snow} & \textbf{mIoU} \\
\cmidrule(lr{0.5em}){1-1} \cmidrule(lr{0.5em}){2-5} \cmidrule(lr{0.5em}){6-6}
PointDR~\cite{xiao20233d} & 26.2 & 30.1 & 50.1 & \underline{43.2} & 34.7 \\
DGLSS~\cite{kim2023single} & \textbf{32.8} & \underline{34.8} & \underline{54.9} & 39.8 & \underline{39.2} \\
SJ+LPD~\cite{park2024rethinking} & 25.4 & 30.6 & 38.2 & 40.6 & 38.3 \\
\rowcolor{gray!10}
\textbf{NTN (Ours)} & \underline{31.8} & \textbf{38.7} & \textbf{58.4} & \textbf{50.4} & \textbf{46.2} \\
\bottomrule
\end{tabular}
\end{adjustbox}
\caption{Performance comparison of different methods under varying weather conditions on SemanticPOSS-to-SemanticSTF benchmark. \textbf{Bold} indicates the best mIoU and \underline{underlined} indicates the second-best performance.}
\label{tab:weather_performance_poss}
\end{table}
%%%%%%%%%%%%%%%%%%%%%%%%%% SEMANTICPOSS 2 SEMANTICSTF WEATHER %%%%%%%%%%%%%%%%%%%%%%%%%%
%%%%%%%%%%%%%%%%%%%%%%%%%% SEMANTICPOSS 2 SEMANTICSTF WEATHER %%%%%%%%%%%%%%%%%%%%%%%%%%

\begin{figure*}[t!]
    \centering
    \includegraphics[width=0.95\textwidth]{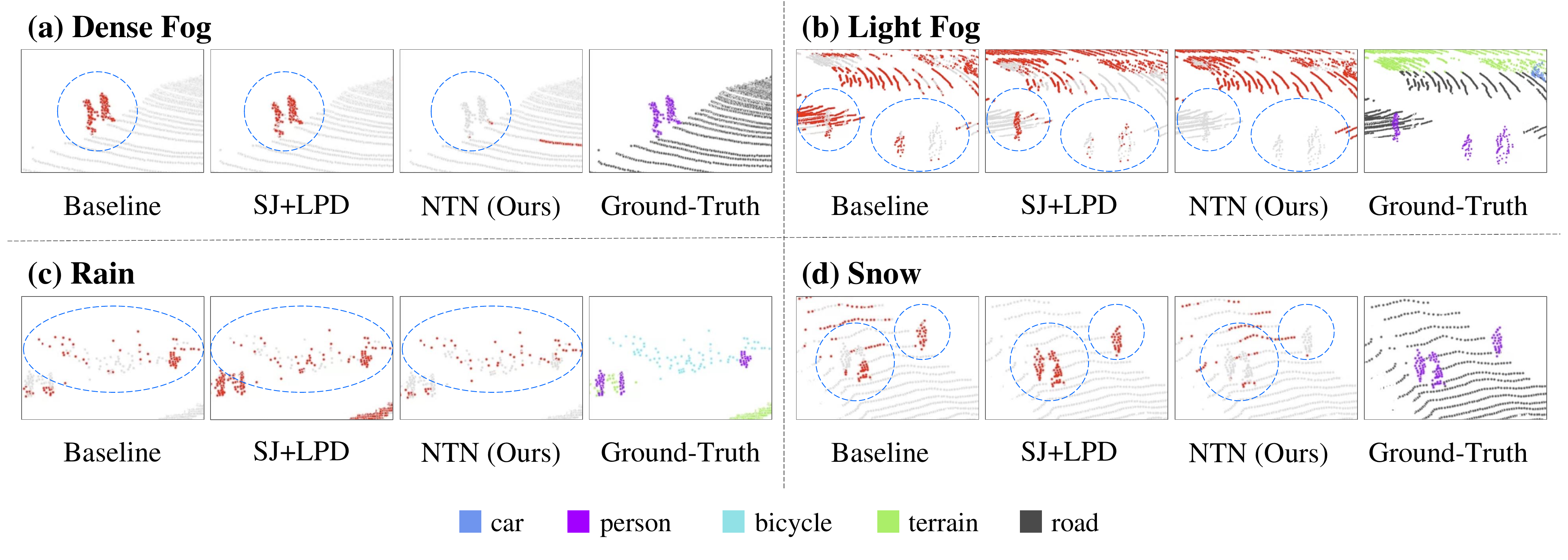}
    \caption{Qualitative results of our method on \textit{validation set} of SemanticSTF. All models are trained on the \textit{train set} of SemanticPOSS. Gray points indicate correct predictions, red points highlight errors, and ground-truth is shown with color-coded labels. Dashed circles highlight the predictions of \textit{things} classes. Best viewed in color.}
    \label{fig:quali_poss}
    %\vspace{-0.5em}
\end{figure*}

\section{Qualitative Results on the SemanticPOSS-to-SemanticSTF benchmark}
\label{sec:quali_poss}

Fig.~\ref{fig:quali_poss} shows qualitative results of our method and baselines on the SemanticSTF validation set, using SemanticPOSS as the source domain. Correct predictions are marked in gray, errors are in red, and ground-truth classes are color-coded. Our method demonstrates significant improvements over the original MinkowskiNet and SJ+LPD in segmenting the \emph{things} category, particularly for classes such as  \texttt{person} and \texttt{bicycle}. In scenarios (a), (b), and (d), our method accurately segments the \texttt{person} class, while others fail to recognize it or generate incomplete segmentation results. These results align with our quantitative analysis, which reveals a 15.5 IoU improvement for the \texttt{person} class. Additionally, in the rain scenario (c), while the segmentation performance of \texttt{bicycle} is not yet optimal, our method still surpasses existing methods, providing better segmentation results. Accurate segmentation of these \emph{things} classes is highly significant, as they directly impact the safety and reliability of autonomous driving.

%%% ================================================================================
%%% ================================================================================
\vspace{1em}
\section{Experiment on SynLiDAR}
\label{sec:synlidar}

%%%%%%%%%%%%%%%%%%%%%%%%%% SYNLIDAR 2 SEMANTICSTF %%%%%%%%%%%%%%%%%%%%%%%%%%
%%%%%%%%%%%%%%%%%%%%%%%%%% SYNLIDAR 2 SEMANTICSTF %%%%%%%%%%%%%%%%%%%%%%%%%%
\begin{table*}[t!]
\centering
\renewcommand{\arraystretch}{1.3}
\begin{adjustbox}{max width=\textwidth}
\arrayrulecolor{black} % black line
\begin{tabular}{l|c c c c c c c c c c c c c c c c c c c | c} % Method와 mIoU 앞뒤로 수직선 추가
\specialrule{1.3pt}{0pt}{3pt} % 두꺼운 윗줄
\textbf{Method} \rule{0pt}{15pt} & car & bi.cle & mt.cle & truck & oth-v. & pers. & bi.clst & mt.clst & road & parki. & sidew. & othe.g. & build. & fence & veget. & trunk & terra. & pole & traf. & \textbf{mIoU} \\
\cmidrule(lr{0.5em}){1-1} \cmidrule(lr{0.5em}){2-20} \cmidrule(lr{0.5em}){21-21} % 줄 단위로 수직선 유지
Oracle & 89.4 & 42.1 & 0.0 & 59.9 & 61.2 & 69.6 & 39.0 & 0.0 & 82.2 & 21.5 & 58.2 & 45.6 & 86.1 & 63.6 & 80.2 & 52.0 & 77.6 & 50.1 & 61.7 & 54.7 \\
\cmidrule(lr{0.5em}){1-1} \cmidrule(lr{0.5em}){2-20} \cmidrule(lr{0.5em}){21-21}
\arrayrulecolor{black} 
Source-only & 27.1 & 3.0 & 0.6 & 15.8 & 0.1 & 25.2 & 1.8 & 5.6 & 23.9 & 0.3 & 14.6 & 0.6 & 36.3 & 19.9 & 37.9 & 17.9 & 41.8 & 9.5 & 2.3 & 15.0 \\
Dropout~\cite{dropout} & 28.0 & 3.0 & 1.4 & 9.6 & 0.0 & 17.1 & 0.8 & 0.7 & 34.2 & 6.8 & 30.5 & 1.1 & 35.5 & 19.1 & 42.3 & 17.6 & 36.0 & 14.0 & 2.8 & 15.2 \\
Perturbation & 27.1 & 2.3 & 2.3 & 16.0 & 0.1 & 23.7 & 1.2 & 4.0 & 27.0 & 3.6 & 16.2 & 0.8 & 29.2 & 16.7 & 35.3 & 18.3 & 17.9 & 5.1 & 2.4 & 15.2 \\
PolarMix~\cite{xiao2022polarmix} & 39.2 & 1.1 & 2.2 & 8.3 & 1.5 & 17.8 & 0.8 & 0.7 & 23.3 & 1.3 & 17.5 & 0.4 & 45.2 & 24.8 & 46.2 & 20.1 & 38.7 & 10.9 & 0.6 & 15.7 \\
MMD~\cite{li2018domain} & 25.5 & 2.3 & 2.1 & 13.2 & 0.7 & 22.1 & 1.4 & 7.5 & 30.8 & 0.4 & 17.6 & 0.4 & 30.9 & 19.7 & 37.6 & 19.3 & 43.5 & 9.9 & 2.6 & 15.1 \\
PCL~\cite{yao2022pcl} & 30.9 & 0.8 & 1.4 & 10.0 & 0.4 & 23.3 & 4.0 & 7.9 & 28.5 & 1.3 & 17.7 & 1.2 & 39.4 & 18.5 & 40.0 & 18.0 & 38.6 & 12.1 & 2.3 & 15.5 \\
PointDR~\cite{xiao20233d} & 37.8 & 2.5 & 2.4 & 23.6 & 0.1 & 26.3 & 2.2 & 7.7 & 27.9 & 7.7 & 17.5 & 0.5 & 47.6 & 25.3 & 45.7 & 21.0 & 37.5 & 17.9 & 5.5 & 18.5 \\
DGLSS~\cite{kim2023single} & 47.9 & 2.9 & 3.4 & 17.4 & 1.1 & 28.0 & 2.4 & 7.3 & 28.8 & 10.2 & 18.1 & 0.2 & 48.9 & 25.3 & 46.5 & 21.4 & 45.2 & 17.9 & 4.9 & 19.8 \\
UniMix~\cite{zhao2024unimix} & 65.4 & 0.1 & 3.9 & 16.9 & 5.3 & 32.3 & 2.0 & 19.3 & 52.1 & 5.0 & 27.3 & 3.0 & 49.4 & 20.3 & 58.5 & 22.7 & 23.2 & 26.1 & 20.9 & 23.4 \\
DGUIL~\cite{he2024domain} & 43.3 & 2.8 & 2.6 & 23.2 & 3.2 & 31.3 & 2.5 & 4.4 & 34.3 & 9.2 & 17.9 & 0.3 & 57.1 & 27.6 & 50.0 & 24.2 & 41.5 & 19.0 & 6.1 & 21.1 \\
\arrayrulecolor{lightgray} 
\cmidrule(lr{1.em}){1-1} \cmidrule(lr{1.em}){2-20} \cmidrule(lr{1.em}){21-21}
SJ+LPD~\cite{park2024rethinking} & 39.0 & 2.5 & 2.5 & 22.3 & 0.3 & 27.0 & 1.8 & 4.0 & 36.1 & 10.3 & 19.0 & 1.0 & 50.6 & 24.5 & 45.1 & 23.2 & 34.1 & 21.9 & 7.2 & 19.6 \\
\rowcolor{gray!10}
\textbf{NTN (Ours)} & 48.4 & 1.5 & 2.4 & 19.4 & 0.2 & 29.1 & 3.2 & 8.9 & 43.5 & 6.7 & 20.5 & 0.0 & 52.2 & 30.1 & 49.8 & 20.0 & 32.9 & 24.7 & 7.5 & 21.1 \\[-0.3em]
\rowcolor{gray!10}
\textbf{\(\uparrow\) to SJ+LPD} & 
{\footnotesize \textcolor{red}{(+9.4)}} & 
{\footnotesize \textcolor{red}{(-1.0)}} & 
{\footnotesize \textcolor{red}{(-0.1)}} & 
{\footnotesize \textcolor{red}{(-2.9)}} & 
{\footnotesize \textcolor{red}{(-0.1)}} & 
{\footnotesize \textcolor{red}{(+2.1)}} & 
{\footnotesize \textcolor{red}{(+1.4)}} & 
{\footnotesize \textcolor{red}{(+4.9)}} & 
{\footnotesize \textcolor[rgb]{0,0.5,0}{(+7.4)}} & 
{\footnotesize \textcolor[rgb]{0,0.5,0}{(-3.6)}} & 
{\footnotesize \textcolor[rgb]{0,0.5,0}{(+1.5)}} & 
{\footnotesize \textcolor[rgb]{0,0.5,0}{(-1.0)}} & 
{\footnotesize \textcolor[rgb]{0,0.5,0}{(+1.6)}} & 
{\footnotesize \textcolor{red}{(+5.6)}} & 
{\footnotesize \textcolor[rgb]{0,0.5,0}{(+4.7)}} & 
{\footnotesize \textcolor[rgb]{0,0.5,0}{(-3.2)}} & 
{\footnotesize \textcolor[rgb]{0,0.5,0}{(-1.2)}} & 
{\footnotesize \textcolor{red}{(+2.8)}} & 
{\footnotesize \textcolor{red}{(+0.3)}} & 
\textcolor{red}{\textbf{(+1.5)}} \\
\arrayrulecolor{black}
\specialrule{1.3pt}{3pt}{0pt} % 두꺼운 아랫줄
\end{tabular}
\end{adjustbox}
\caption{Comparison of methods on the SynLiDAR-to-SemanticSTF benchmark. Performance improvements of our method over SJ+LPD~\cite{park2024rethinking} are shown in the last row, with red text indicating \emph{things} classes increments and green text for \emph{stuff}.}
\label{tab:main_exp_syn2stf}
\end{table*}
%%%%%%%%%%%%%%%%%%%%%%%%%% SYNLIDAR 2 SEMANTICSTF %%%%%%%%%%%%%%%%%%%%%%%%%%
%%%%%%%%%%%%%%%%%%%%%%%%%% SYNLIDAR 2 SEMANTICSTF %%%%%%%%%%%%%%%%%%%%%%%%%%

%%%%%%%%%%%??????????????
We conduct additional experiments on the SynLiDAR-to-SemanticSTF benchmark. As shown in Table~\ref{tab:main_exp_syn2stf}, our method improves the overall performance by 1.5 mIoU compared to the SJ+LPD model~\cite{park2024rethinking}. Although our proposed methods provide slight performance benefits, the improvements are not as significant as those observed in other benchmarks. 
SynLiDAR-to-SemanticSTF entangles both (1) the domain gap between synthetic and real data and (2) weather corruption. This is why our proposed methods provide marginal performance gain.
Nevertheless, Table ~\ref{tab:main_exp_syn2stf} shows FB and BFD outperform the previous method, SJ+LPD~\cite{park2024rethinking}, implying its effectiveness in representing real adverse weather.
Note that UniMix~\cite{zhao2024unimix} and DGUIL~\cite{he2024domain} have not been reproduced due to the lack of publicly available code.

% Our method uses real clean data as the source and targets real adverse data. In contrast, SynLiDAR-to-SemanticSTF involves using synthetic clean data as the source to target real adverse data. There are differences in LiDAR point density and object appearance between synthetic and real data. 
% Therefore, in addition to weather-induced corruption, there is also a domain difference between synthetic and real data. As a result, our proposed method does not perfectly align with the task on SynLiDAR-to-SemanticSTF benchmark, which may explain why the performance improvement is not as large.

%%%%%%%%%%%%%%%%%%%%%%%%%%%%%%%%%%%%%%%%%%%%%%%%%%%

\vspace{1em}
\section{Discussions about Methods}
\label{sec:analysis}

In this section, we discuss both the effectiveness and limitations of our proposed modules on the SemanticKITTI-to-SemanticSTF benchmark.

%%%% ====
\begin{figure}[bt!]
    \centering
    \includegraphics[width=\linewidth]{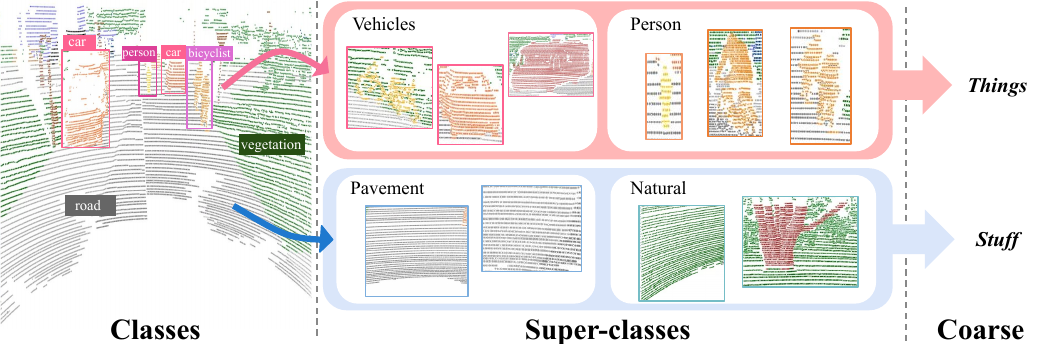}
    \caption{Examples of the three levels of category classification: classes, superclasses, and coarse, in increasing order of granularity.}
    \label{fig:class_type_def}
\end{figure}
%%%% ====

\subsection{Analysis on Feature Binding}
\noindent\textbf{Multi-Level Categories.}
We divide categories into three levels, as shown in Fig.~\ref{fig:class_type_def}. Classes are the smallest units, superclasses group semantically similar classes, and coarse categories consist of \emph{things} and \emph{stuff}. We perform a detailed analysis of these three levels using confusion matrices and qualitative results.

%%%% ====
\begin{figure}[tb!]
    \centering
    \includegraphics[width=\linewidth]{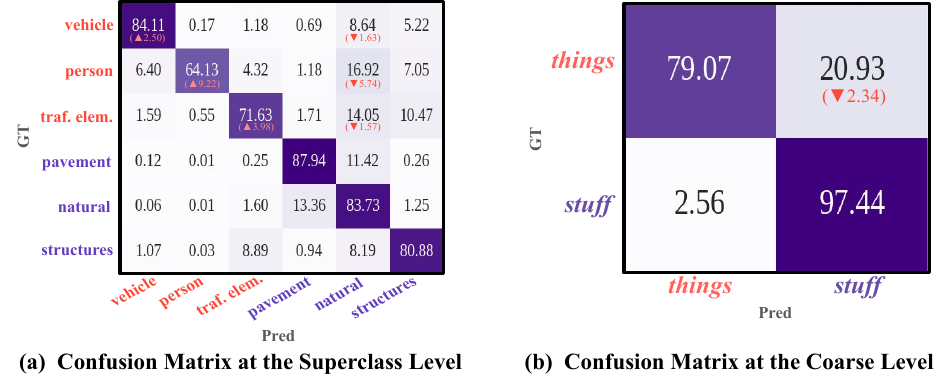}
    \caption{Confusion matrices for (a) superclasses and (b) \textit{things} and \textit{stuff} after adding Feature Binding. The numbers in brackets indicate the increase or decrease compared to SJ+LPD~\cite{park2024rethinking}. FB effectively reduces misclassification between \textit{things} and \textit{stuff}.}
    \label{fig:conf_fb}
\end{figure}
%%%% ====

\vspace{0.5em}
% \subsection{Analysis on Feature Binding}
\noindent\textbf{Effect of FB for Relieving Confusion.}
As mentioned in Sec.~\ref{sec:rationale}, Feature Binding is a module designed to prevent \emph{things} from being misclassified as \emph{stuff} when their semantic information is partially degraded due to corruption caused by adverse weather.
To examine the effect of FB, we conduct an analysis at the superclass level and the coarse level, as shown in Fig.~\ref{fig:conf_fb}. When we compare the performance with the previous method~\cite{park2024rethinking}, we find that misclassification ratio of \emph{things} as \emph{stuff} decreases by 2.34\%.
At the superclass level, we observe that the tendency to misclassify \emph{things} as \emph{natural} categories such as \texttt{vegetation} is greatly reduced. In particular, for the \emph{person} category, true positive prediction ratio increases by 9.22\% compared to before, and misclassification ratio as the \emph{natural} category decreases by 5.74\%.
Thus, we confirm that FB helps to learn the semantic differences between \emph{things} and \emph{stuff} by continuously providing hierarchical semantic information.

%%%%%%% only FB, only BFD에 대한 class-wise conf matrix %%%%%%%
%%%%%%%%%%%%%%% 필요할 경우에만 쓰면 될 것 같음 %%%%%%%%%%%%%%%%%%
% \begin{figure}[tb!]
%     \centering
%     \includegraphics[width=\linewidth]{figures/full_conf_matrix_ours_only_FB.pdf}
%     \caption{full conf matrix Ours (only FB)}
%     \label{fig:conf_full}
% \end{figure}

% \begin{figure}[tb!]
%     \centering
%     \includegraphics[width=\linewidth]{figures/full_conf_matrix_ours_only_BFD.pdf}
%     \caption{full conf matrix Ours (only BFD)}
%     \label{fig:conf_full}
% \end{figure}
%%%%%%%%%%%%%%%%%%%%%%%%%%%%%%%%%%%%%%%%%%%%%%%%%%%%%%%%%%%%%
%%%%%%%%%%%%%%%%%%%%%%%%%%%%%%%%%%%%%%%%%%%%%%%%%%%%%%%%%%%%%

%%%% ====

%%%% ====
\begin{figure}[tb!]
    \centering
    \includegraphics[width=0.82\linewidth]{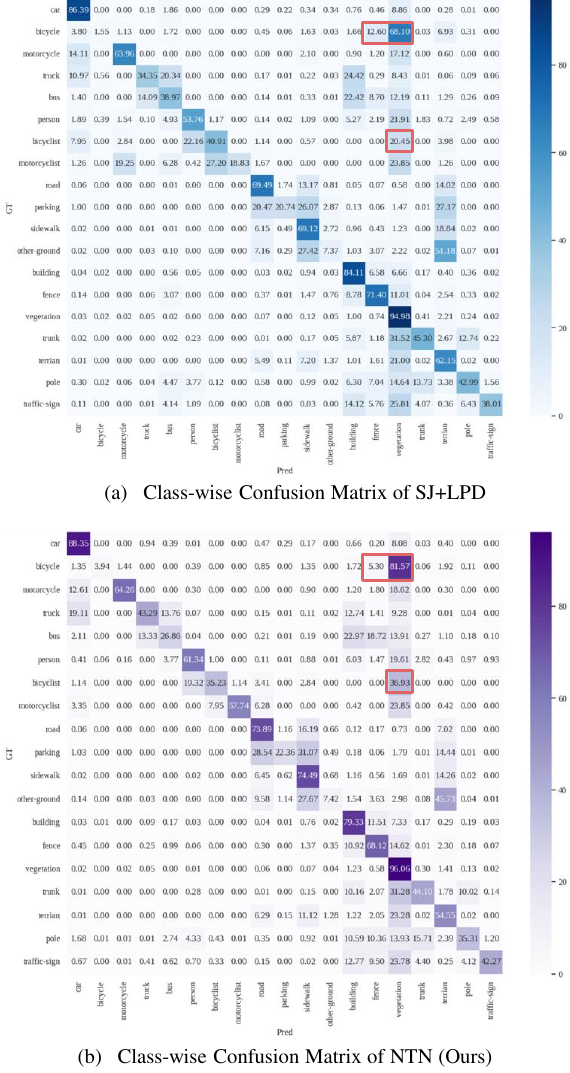}
    \caption{Class-wise confusion matrices of (a) SJ+LPD~\cite{park2024rethinking} and (b) Ours on SemanticKITTI-to-SemanticSTF benchmark.}
    \label{fig:full_confusion_matrix_supp}
\end{figure}
%%%% ====

\begin{figure}[tb!]
    \centering
    \includegraphics[width=\linewidth]{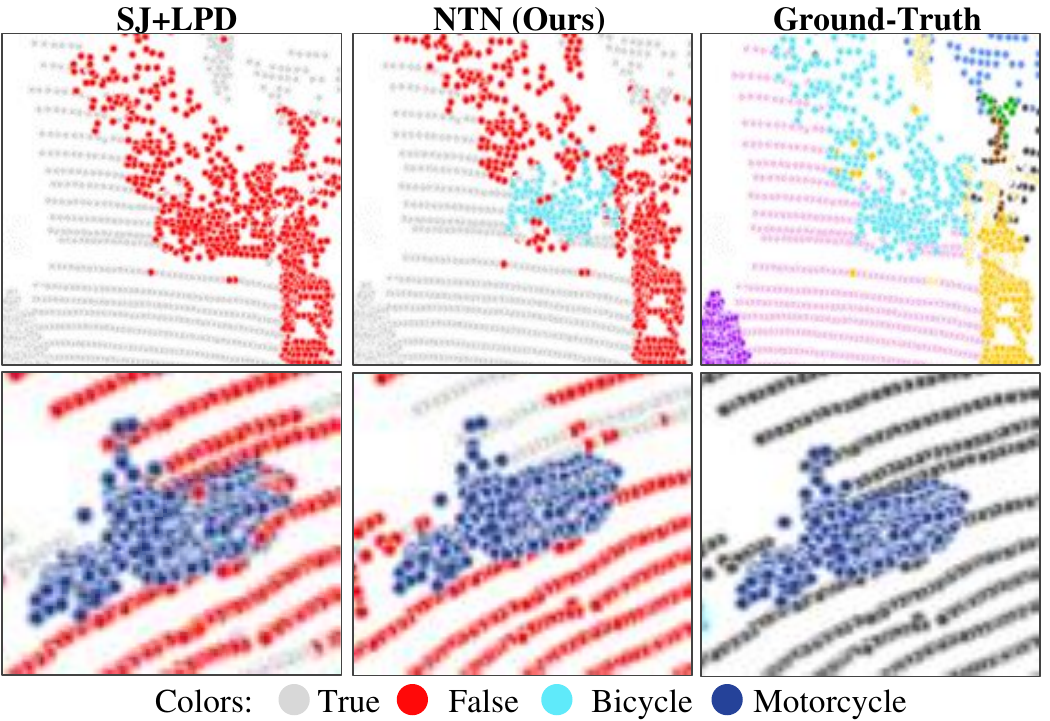}
    \caption{Qualitative results of NTN (FB + BFD). Gray is true prediction, and Red is false prediction. Other colors mean true predictions for specific classes.}
    \label{fig:scan8_11}
\end{figure}

\begin{table}[tb!]
\centering
\renewcommand{\arraystretch}{1.3} % 줄 간격 조정
\arrayrulecolor{black}
\begin{adjustbox}{max width=0.95\columnwidth}
\begin{tabular}{c|c|ccc}
\specialrule{1.2pt}{0pt}{3pt}
\multicolumn{2}{c|}{\textbf{Feature Distillation Types}} & \multicolumn{3}{c}{\textbf{Performance}} \\
\textit{$\mathcal{B}$-all.} & \textit{$\mathcal{B}$-wise.} (Ours) & \textbf{Things} & \textbf{Stuff} & \textbf{mIoU} \\
\cmidrule(lr{0.5em}){1-1} \cmidrule(lr{0.5em}){2-2} \cmidrule(lr{0.5em}){3-4} \cmidrule(lr{0.5em}){5-5}
\checkmark & \xmark & 35.3 & 43.7 & 38.9 \\
\xmark & \checkmark & \textbf{36.2}  & 42.5 & 38.9  \\
 \arrayrulecolor{black}
\specialrule{1.2pt}{0pt}{3pt}
\end{tabular}
\end{adjustbox}
\caption{Comparison for feature distillation settings between average on all beams and seperate beam. Experiments are done on SemanticKITTI-to-SemanticSTF benchmark.}
\label{tab:comparison_beamavg_beamwise}
\vspace{-0.8em}
\end{table}

\vspace{0.5em}
% \section{Analysis on Beam-wise Feature Distillation.} 
\subsection{Analysis on Beam-wise Feature Distillation.}
\noindent\textbf{Challenges of BFD with Disrupted Point Patterns.}
\label{sec:analysis_bfd}
As in Fig.~\ref{fig:full_confusion_matrix_supp}, misprediction ratio from things-to-stuff slightly increases in \texttt{bicycle} and \texttt{bicyclist} classes.
This is because BFD tends to effectively utilize the point patterns before point drop, as mentioned in Sec.~\ref{sec:rationale}.

The learning mechanism of BFD enables the LSS model to perform better on objects where the clean branch's point pattern is well preserved during test time. 
This explains why our method struggles with classes like \texttt{bicycle} and \texttt{bicyclist}, which inherently have few points and are prone to severe point pattern corruptions. In such cases, BFD struggles to utilize informations of local point pattern from clean data branch.
Conversely, significant performance improvements are observed for classes like \texttt{motorcycle} and \texttt{motorcyclist}, which have clearer shapes.
As shown in Fig.~\ref{fig:scan8_11}, the method with BFD demonstrates superior predictions for bicycles with reasonably preserved point patterns, as observed in row 1. 
Also, the \texttt{motorcycle} in row 2, which maintains its point pattern, achieves improved performance.
In summary, confusion on classes like \texttt{bicycle} arises as a side effect of improving performance by preserving \emph{things} object information without bias toward weather-corrupted data.

% Abaltion with class mean
\vspace{0.5em}
\noindent\textbf{Ablation Study on Types of Feature Distillation.}
\label{sec:non-beam-wise}
As shown in Table~\ref{tab:comparison_beamavg_beamwise}, distillation after averaging across all beams (\textit{$\mathcal{B}$-all.}) achieves the same mIoU as our method, which performs beam-wise averaging before distillation (\textit{$\mathcal{B}$-wise.}). 
However, our method outperforms \textit{$\mathcal{B}$-all.} in \emph{things} objects. This demonstrates that defining beam-wise local regions and performing distillation accordingly better compensates for information loss caused by point missing in \emph{things} objects.

\begin{figure}[tb!]
    \centering
    \includegraphics[width=\linewidth]{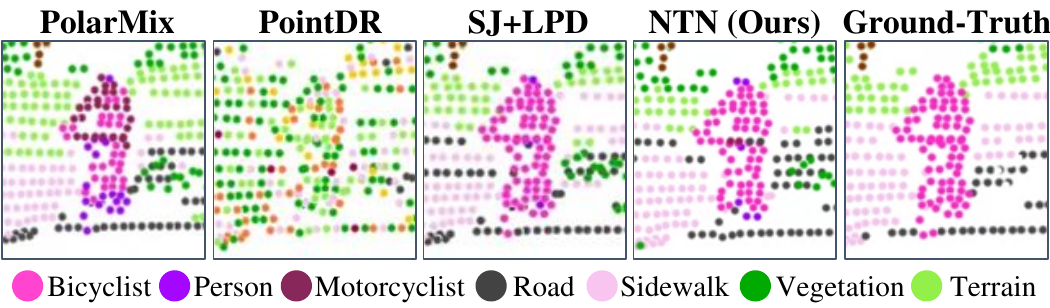}
    \caption{Qualitative results on \texttt{bicyclist} class for PolarMix~\cite{xiao2022polarmix}, PointDR~\cite{xiao20233d}, SJ+LPD~\cite{park2024rethinking} and Ours.}
    \label{fig:scan155}
\end{figure}

\begin{table*}[tb!]
\centering
\resizebox{\textwidth}{!}{%
\renewcommand{\arraystretch}{0.75}
\begin{tabular}
%\begin{tblr}
% {
%     colspec = {l|ccc||c|ccc||ccc|ccc},
%     %column{4} = {pink!3},
%   }
{l|ccc||c|ccc|ccc|ccc}
\specialrule{1.2pt}{0pt}{3pt}
% ---------------------------
\multicolumn{1}{>{\columncolor{pink!20}}c|}{Dataset}
& \multicolumn{3}{>{\columncolor{pink!20}}c||}{\textbf{Sem.KITTI}; w. Other Sim.}
& \multicolumn{1}{>{\columncolor{cyan!25}}c|}{Dataset}
& \multicolumn{3}{>{\columncolor{cyan!25}}c|}{\textbf{nuScenes}; real rain}
& \multicolumn{6}{>{\columncolor{cyan!15}}c}{\textbf{nuScenes-C}; snow, fog, wet ground} \\
\cmidrule(lr{0.5em}){1-1}
\cmidrule(lr{0.5em}){2-5}
\cmidrule(lr{0.5em}){6-8}
\cmidrule(lr{0.5em}){9-14}
% ---------------------------
Method
& \textit{Things} & \textit{Stuff} & \textbf{All}
& Method
& \textit{Things} & \textit{Stuff} & \textbf{All}
& Heavy & Moderate & Light & \textit{Things} & \textit{Stuff} & \textbf{All} \\
\cmidrule(lr{0.5em}){1-1}
\cmidrule(lr{0.5em}){2-3}
\cmidrule(lr{0.5em}){4-4}
\cmidrule(lr{0.5em}){5-5}
\cmidrule(lr{0.5em}){6-7}
\cmidrule(lr{0.5em}){8-8}
\cmidrule(lr{0.5em}){9-11}
\cmidrule(lr{0.5em}){12-13}
\cmidrule(lr{0.5em}){14-14}

% 1) Source Only
Source Only
& 17.7 & 38.9 & 26.6
& \multicolumn{1}{l|}{Source Only}
& 34.2 & 67.0 & 48.6
& 47.5 & 52.5 & 55.1 & 39.0 & 68.0 & 51.7 \\

% 2) SJ+LPD
% \multicolumn{1}{l|}{[\textcolor{RoyalBlue}{A}]+[\textcolor{RoyalBlue}{B}]}
\multicolumn{1}{l|}{\cite{hahner2021fog} + \cite{hahner2022lidar}}
& 22.9 & 42.8 & 31.3
& \multicolumn{1}{l|}{SJ+LPD}
& 33.2 & 66.6 & 47.8
& 49.0 & 53.6 & 55.7 & 40.7 & 68.2 & 52.7 \\[+0.05em]

% \rowcolor{gray!10}
% \arrayrulecolor{gray!50}
% \hline
% \arrayrulecolor{black}
\rowcolor{gray!10}
\textbf{+ NTN (Ours)}
& \textbf{23.6} & \textbf{43.9} & \textbf{32.2}
& \multicolumn{1}{l|}{\textbf{+ NTN (Ours)}}
& \textbf{35.4} & \textbf{67.8} & \textbf{49.6}
& \textbf{48.6} & \textbf{54.0} & \textbf{56.6} & \textbf{53.1} & \textbf{68.5} & \textbf{53.1} \\[-0.15em]

\rowcolor{gray!10}
% \multicolumn{1}{l|}{\footnotesize \textbf{\(\uparrow\) to [\textcolor{RoyalBlue}{A}] + [\textcolor{RoyalBlue}{B]}}
\multicolumn{1}{l|}{\footnotesize \textbf{\(\uparrow\) to \cite{hahner2021fog} + \cite{hahner2022lidar}}}
& {\footnotesize \textcolor{red}{(+0.7)}}
& {\footnotesize \textcolor{gray!150}{(+1.1)}}
& {\footnotesize \textcolor{red}{\textbf{(+0.9)}}}
& \multicolumn{1}{l|}{{\footnotesize \textbf{\(\uparrow\) to SJ+LPD}}}
& {\footnotesize \textcolor{red}{(+2.2)}}
& {\footnotesize \textcolor{gray!150}{(+1.2)}}
& {\footnotesize \textcolor{red}{\textbf{(+1.8)}}}
& {\footnotesize \textcolor{red}{(+2.2)}}
& {\footnotesize \textcolor{red}{(+1.9)}}
& {\footnotesize \textcolor{red}{(+1.7)}}
& {\footnotesize \textcolor{red}{(+12.4)}}
& {\footnotesize \textcolor{gray!150}{(+0.3)}}
& {\footnotesize \textcolor{red}{\textbf{(+0.4)}}} \\
\specialrule{1.2pt}{0pt}{3pt}
%\end{tblr}
\end{tabular}
}
% \vspace{-1em}
\caption{Comparison on various benchmarks and scenarios.}
\label{tab:nuscenes-c}
% \vspace{-0.75em}
\end{table*}

\begin{table*}[tb!]
\centering
\renewcommand{\arraystretch}{0.65} % 줄 간격 조정
\arrayrulecolor{black}
\begin{adjustbox}{max width=0.8\textwidth}
\begin{tabular}{c|ll|c|c}
\specialrule{1.1pt}{0pt}{2pt}
 Method & \multicolumn{2}{c|}{\textbf{Superclass}} & \textbf{mIoU} & \textbf{Total} \\
\cmidrule(lr{0.5em}){1-1} \cmidrule(lr{0.5em}){2-3} \cmidrule(lr{0.5em}){4-4} \cmidrule(lr{0.5em}){5-5}
 \multirow{2}{*}{Coarse} & \multicolumn{2}{l|}{\textbf{Things:} \footnotesize{\texttt{person, vehicle, traffic element}}} & 33.7 \textcolor{red}{\scriptsize (\(\uparrow\)2.4)} & \multirow{2}{*}{\textbf{38.1} \textcolor{red}{\scriptsize (\(\uparrow\)1.8)}}\\
 & \multicolumn{2}{l|}{\textbf{Stuff:} \footnotesize{\texttt{pavement, natural, structure}}} & 44.0 \textcolor{gray}{\scriptsize (\(\uparrow\)0.9)} & \\
 % \arrayrulecolor{lightgray}
 \cmidrule(lr{0.5em}){1-5}
 \arrayrulecolor{black}
  \multirow{6}{*}{GPT-o1} & \multicolumn{2}{l|}{\textbf{Person}: \footnotesize{\texttt{person, bi.clst, mt.clst}}} & 39.1 \textcolor{red}{\scriptsize (\(\uparrow\)13.4)} & \multirow{6}{*}{\textbf{38.3} \textcolor{red}{\scriptsize (\(\uparrow\)2.0)}}\\ % (SJ+LPD)
  & \multicolumn{2}{l|}{\textbf{Vehicle}: \footnotesize{\texttt{car, bi.cle, mt.cle, truck, oth-v.}}} & 33.5 \textcolor{red}{\scriptsize (\(\uparrow\)3.4)} &  \\
  & \multicolumn{2}{l|}{\textbf{Ground:} \footnotesize{\texttt{road, parki., sidew., othe.g.}}} & 30.0 \textcolor{gray}{\scriptsize (\(\downarrow\)0.8)} & \\
  & \multicolumn{2}{l|}{\textbf{Nature:} \footnotesize{\texttt{trunk, veget., terra.}}} & 48.6 \textcolor{gray}{\scriptsize (\(\uparrow\)0.4)} & \\
  & \multicolumn{2}{l|}{\textbf{Construction:} \footnotesize{\texttt{build., fence}}} & 57.3 \textcolor{gray}{\scriptsize (\(\downarrow\)6.8)} & \\
  & \multicolumn{2}{l|}{\textbf{Traffic Object:} \footnotesize{\texttt{pole, traf.}}} & 31.5 \textcolor{gray}{\scriptsize (\(\downarrow\)2.0)} & \\
 \arrayrulecolor{black}
 \cmidrule(lr{0.5em}){1-5}
 \arrayrulecolor{black}
  \multirow{6}{*}{DeepSeek-V3} & \multicolumn{2}{l|}{\textbf{Person}: \footnotesize{\texttt{person, bi.clst, mt.clst}}} & 30.0 \textcolor{red}{\scriptsize (\(\uparrow\)4.4)} & \multirow{6}{*}{\textbf{39.4} \textcolor{red}{\scriptsize (\(\uparrow\)3.1)}}\\ % (SJ+LPD)
  & \multicolumn{2}{l|}{\textbf{Vehicle}: \footnotesize{\texttt{car, bi.cle, mt.cle, truck, oth-v.}}} & 37.3 \textcolor{red}{\scriptsize (\(\uparrow\)7.2)} &  \\
  & \multicolumn{2}{l|}{\textbf{Road-infra.:} \footnotesize{\texttt{road, parki., sidew., othe.g.}}} & 32.1 \textcolor{gray}{\scriptsize (\(\uparrow\)1.3)} & \\
  & \multicolumn{2}{l|}{\textbf{Nature:} \footnotesize{\texttt{veget., terra.}}} & 56.6 \textcolor{gray}{\scriptsize (\(\uparrow\)2.9)} & \\
  & \multicolumn{2}{l|}{\textbf{Man-made:} \footnotesize{\texttt{build., fence, pole, traf.}}} & 49.1 \textcolor{gray}{\scriptsize (\(\uparrow\)0.4)} & \\
  & \multicolumn{2}{l|}{\textbf{Miscell.:} \footnotesize{\texttt{trunk}}} & 35.1 \textcolor{gray}{\scriptsize (\(\downarrow\)2.1)} \\
 \arrayrulecolor{black}
\specialrule{1.2pt}{0pt}{2pt}
\end{tabular}
\end{adjustbox}
% \vspace{-1em}
\caption{Comparison of different superclasses on SemanticKITTI-to-SemanticSTF (improvements over SJ+LPD in brackets).}
\label{tab:superclass_more}
% \vspace{-1.5em}
\end{table*}

% \vspace{-0.5em}
\section{Analysis on mIoU Drop \texttt{Bicyclist} \& \texttt{Fence} Classes.}
\label{sec:bicyclist_miou}
\noindent\textbf{Bicyclist.} We found out that both the predictions from SJ+LPD~\cite{park2024rethinking} and our method accurately detected the bicyclist within a single scan. 
As shown in the third and fourth visualizations of Fig.~\ref{fig:scan155}, the performance difference between SJ+LPD and our method is minimal, amounting to only a few points. Furthermore, the mispredicted points were all classified as a person, indicating that the slight performance drop does not significantly impact safety-critical predictions.

We verified the results for this object using PointDR and PolarMix, which are reproducible due to publicly available codes.
As illustrated in the first and second visualizations of Fig.~\ref{fig:scan155}, both PolarMix~\cite{xiao2022polarmix} and PointDR~\cite{xiao20233d} failed to provide accurate predictions for most bicycle points.
In PolarMix, some points are predicted as \texttt{bicyclist}, but many are misclassified as \texttt{motorcyclist}. As \texttt{motorcyclist} generally moves faster than \texttt{bicyclist} or \texttt{person}, such mispredictions could lead to significant risks in ensuring safe driving.
For PointDR, none of the points are predicted as \texttt{person}, \texttt{bicyclist}, or \texttt{motorcyclist}.
This demonstrates that previous methods fail to predict even within the \texttt{person} superclass, highlighting the effectiveness of our method in safety-critical driving scenarios.

\noindent\textbf{Fence.} Unlike other thin traffic elements, \texttt{fence} varies from thin to large structures. Due to this structural diversity, large fences receive confusing signals from the distinct-shaped superclass by FB. 
As a result, \texttt{fence} had lower IoU and increased confusion with \texttt{building} and \texttt{vegetation} as in Fig.~\ref{fig:full_confusion_matrix_supp}.

\section{Performance Degradation of \texttt{Car} Class on SemanticPOSS$\rightarrow$SemanticSTF.}
\label{sec:car_poss2stf}
The performance gap arises from differences in \texttt{car} class annotation between SemanticKITTI and SemanticPOSS. 
SemanticKITTI-trained model benefits from FB by learning diverse vehicle classes separately, thus effective in SemanticSTF. In contrast, SemanticPOSS groups \texttt{car}, \texttt{bus}, and \texttt{truck} into one class, limiting feature diversity and lowering performance. 
For optimal performance of FB, pre-defined classes should be as fine-grained as possible.

%-------------------------------------------------------------------------

\section{Related Works on Subclass- or Prototype-based Methods}
\label{sec:related_works_more}
Subclass- or Prototype-based regularization has frequently been employed to tackle label-efficiency problems, where limited or imbalanced annotations lead to suboptimal segmentation. Pixel-to-Prototype Contrast~\cite{pixel2proto_cvpr2022} addresses weakly supervised semantic segmentation by aligning pixel embeddings with class prototypes, refining noisy pseudo masks generated from image-level labels. Similarly, Prototypical Contrastive Network~\cite{proto_network_wacv2024} focuses on highly imbalanced aerial segmentation by learning a single foreground prototype and pushing away hard-negative background features, thus emphasizing minority classes. Unbiased Subclass Regularization~\cite{unbiased_subclass_cvpr2022} aims to mitigate semi-supervised class imbalance by splitting overrepresented classes into smaller clusters, forming class-balanced subclasses. Despite sharing the general goal of improving segmentation via prototype or class-based groupings, our work diverges in both methodology and application. Instead of subdividing a single class or maintaining per-class prototypes, we merge multiple classes into higher-level \emph{superclasses} (e.g., \emph{things} vs. \emph{stuff}), which better addresses the broad semantic gap under adverse weather in LiDAR data. Moreover, prior methods largely concentrate on 2D image tasks or semi-/weakly supervised settings, whereas we focus on single-domain generalization for 3D LiDAR segmentation, emphasizing robustness against severe weather-induced corruptions.

\section{Additional Experiments on Different Weather Simulations.}
\label{sec:weather_simulation_more}
We used simulation methods \cite{hahner2021fog, hahner2022lidar} to generate occlusion-induced points, improving performance (Table~\ref{tab:nuscenes-c}) and confirming robustness to other weather simulations. 
Similar to Table~\ref{fig:overview}, the performance gain was higher in \textit{things} (+0.7 mIoU), highlighting our method's safety benefits.

\section{Additional Experiments on Various Datasets}
\label{sec:experiments_more}
We tested our method on two settings: (1) nuScenes \emph{clean train set}-to-\emph{rainy splits}, and (2) nuScenes \emph{train set}-to-\emph{nuScenes-C}. 
As shown in Table~\ref{tab:nuscenes-c}, our method consistently surpasses SJ+LPD~\cite{park2024rethinking} across diverse datasets, particularly for \textit{things}.

\section{Additional Experiments for Different Superclass.}
\label{sec:superclass_more}
We evaluated the impact of superclass selection on generalizability by: (1) separating superclasses into \textit{things} and \textit{stuff}, (2) using GPT-o1, and (3) Deepseek-V3 recommendations. 
For (2) and (3), we provide COCO and ImageNet supercategory examples. 
Table~\ref{tab:superclass_more} shows that even (1) helps FB reduce \emph{things}$\rightarrow$\emph{stuff} mispredictions and improves performance, while finer superclasses in (2) and (3) achieve higher gains. 
This demonstrates well-chosen superclasses can boost performance and highlight the scalability of our method. 
While our method consistently improves \textit{things}, future work is needed for automatic superclass selection.
%-------------------------------------------------------------------------

% \vspace{0.5em}
\section{Additional Visual Comparisons Between Clean and Corrupted Data}
\label{sec:examples}
% We provide additional comparisons between clean and corrupted data, as mentioned in 

\begin{figure*}[t!]
    \centering
    \includegraphics[width=\textwidth]{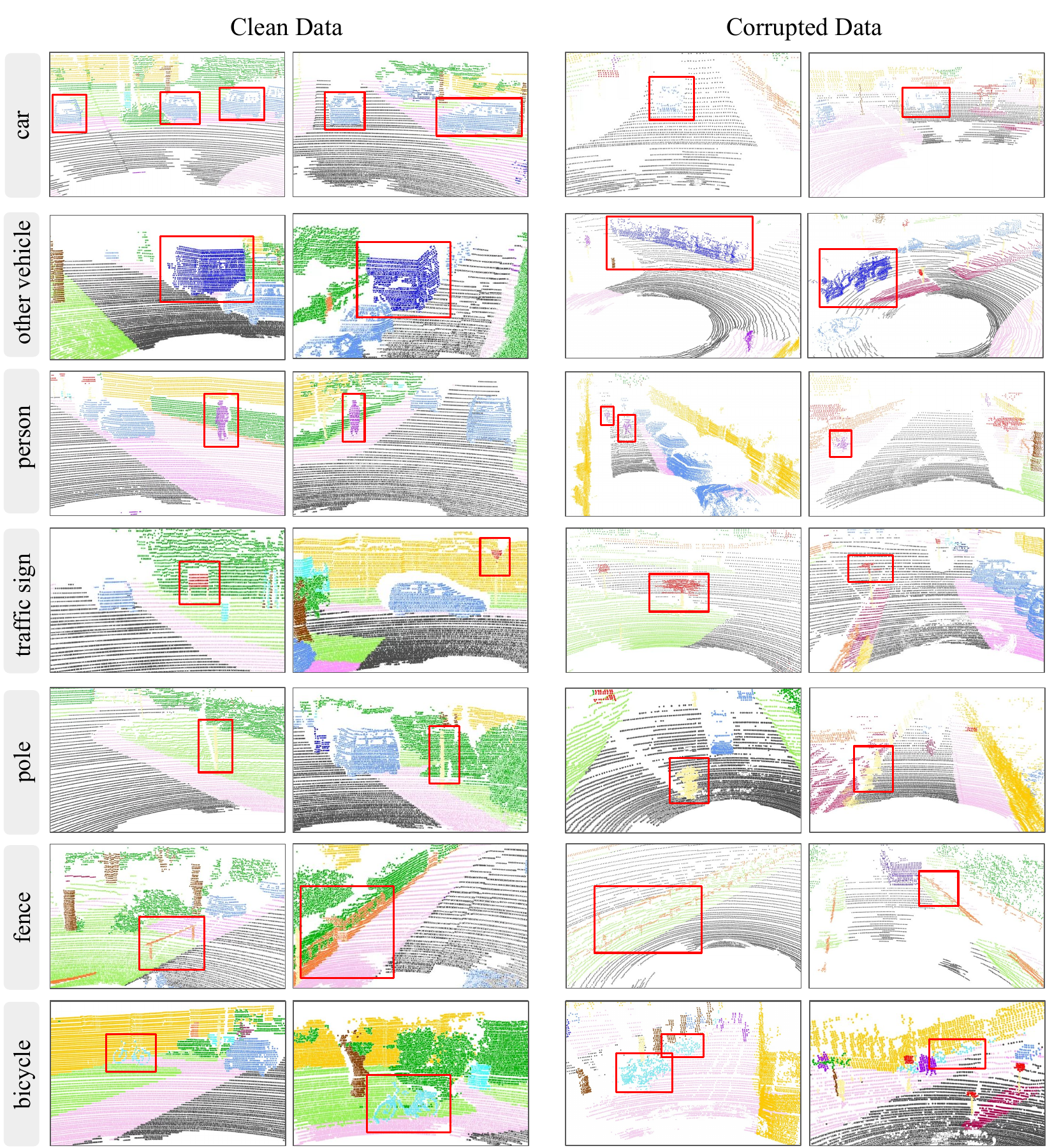}
    \caption{Examples of clean and corrupted data for various \emph{things} classes. In clean weather, objects have well-constructed shapes with dense point clouds. In contrast, objects in adverse weather have blurred shapes with significant point loss.}
    \label{fig:clean_vs_corrup}
    \vspace{5em}
\end{figure*}

% \noindent\textbf{Additional Comparisons between Clean and Corrupted Data.}
We provide additional comparisons between clean and corrupted data, as shown in Fig.~\ref{fig:clean_vs_corrup}. It illustrates point patterns of \emph{things} objects in clean and weather-induced corrupted data.
As mentioned in Sec. \textcolor{cyan}{1} and Sec. \textcolor{cyan}{3} in the main paper, \emph{things} objects in clean weather have well-defined shapes and smooth boundaries, whereas in adverse weather, they have blurred shapes and irregular boundaries.
This proves that \emph{things} classes are more vulnerable to such noise or point loss, making accurate predictions more difficult.

%%%% ====
\vspace{2em}
%%%% ====

%%%% ====
\begin{figure*}[t!]
    \centering
    \includegraphics[width=\textwidth]{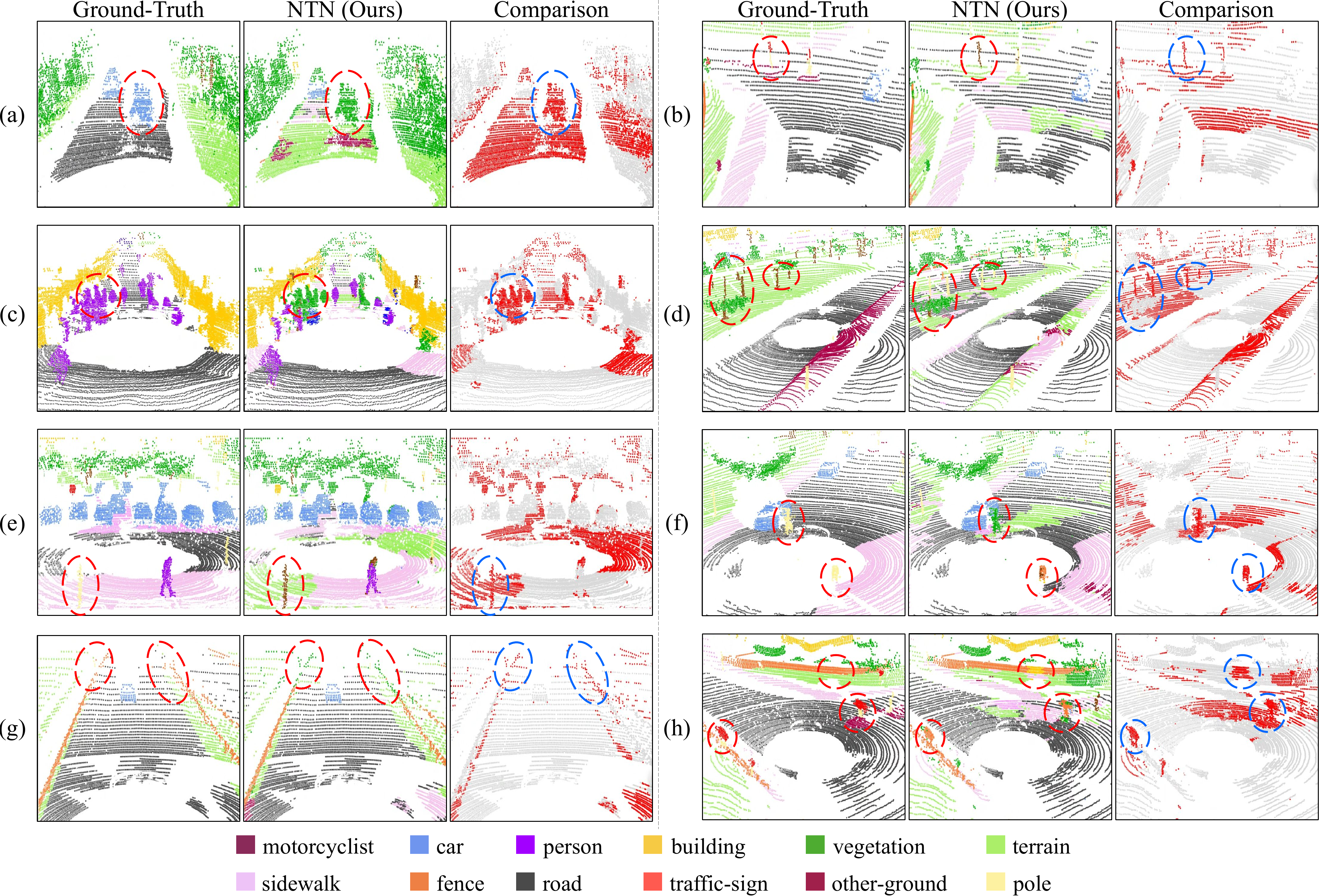}
    \caption{Qualitative results of failure cases on the \textit{validation set} of SemanticSTF. Ground-truth and our prediction results are shown with color-coded labels. For comparison, gray points indicate correct predictions and red points highlight errors.}
    \label{fig:failure_qual}
\end{figure*}
%%%% ====

\section{Failure Cases}
\label{sec:failure_case}
Fig.~\ref{fig:failure_qual} illustrates cases where our method fails in prediction.
Errors remain, particularly at very close distances to vehicles, as shown in (a) and (e).
As demonstrated in (a), performance improvements were limited for the \texttt{car} class due to extreme sparsity from occlusion by droplets.
For the \texttt{person} class, (c) shows yet many incorrect predictions remain in extremely noisy conditions.
Thin objects like \texttt{pole} and \texttt{traffic-sign} are often misclassified as vegetation, as seen in cases (b), (d), (e), (f) and (h).
Errors also occur in classes with varying object sizes, such as fences in examples (g) and (h). As discussed in Sec.~\ref{sec:bicyclist_miou}, this issue arises from defining FB superclasses manually.
{
    \small
    \bibliographystyle{ieeenat_fullname}
    \bibliography{main}
}

\end{document}